\pdfoutput=1

\documentclass[11pt]{article}

\usepackage[dvipsnames]{xcolor}

\usepackage{EMNLP2022}

\usepackage{times}
\usepackage{latexsym}

\usepackage[T1]{fontenc}

\usepackage[utf8]{inputenc}

\usepackage{microtype}

\usepackage[]{bbm}

\usepackage{inconsolata}

\usepackage{graphicx}
\usepackage{multirow}
\usepackage{xspace}

\usepackage[nameinlink]{cleveref}

\newcommand{\aautoref}[1]{\hyperref[#1]{Appendix~\ref*{#1}}}

\usepackage{floatrow}
\newfloatcommand{capbtabbox}{table}[][\FBwidth]

\usepackage{pgfplots}
\pgfplotsset{width=30cm,compat=1.9}

\pgfplotscreateplotcyclelist{mycolorlist}{
Aquamarine,style={fill=.!40}\\%
Bittersweet,style={fill=.!40}\\%
Black,style={fill=.!40}\\%
BlueGreen,style={fill=.!40}\\%
BlueViolet,style={fill=.!40}\\%
BrickRed,style={fill=.!40}\\%
Brown,style={fill=.!40}\\%
BurntOrange,style={fill=.!40}\\%
CadetBlue,style={fill=.!40}\\%
CarnationPink,style={fill=.!40}\\%
Cerulean,style={fill=.!40}\\%
CornflowerBlue,style={fill=.!40}\\%
Dandelion,style={fill=.!40}\\%
DarkOrchid,style={fill=.!40}\\%
Emerald,style={fill=.!40}\\%
ForestGreen,style={fill=.!40}\\%
Fuchsia,style={fill=.!40}\\%
Goldenrod,style={fill=.!40}\\%
OliveGreen,style={fill=.!40}\\%
GreenYellow,style={fill=.!40}\\%
JungleGreen,style={fill=.!40}\\%
Lavender,style={fill=.!40}\\%
LimeGreen,style={fill=.!40}\\%
Magenta,style={fill=.!40}\\%
Mahogany,style={fill=.!40}\\%
Maroon,style={fill=.!40}\\%
Melon,style={fill=.!40}\\%
MidnightBlue,style={fill=.!40}\\%
Mulberry,style={fill=.!40}\\%
NavyBlue,style={fill=.!40}\\%
Gray,style={fill=.!40}\\%
Orange,style={fill=.!40}\\%
OrangeRed,style={fill=.!40}\\%
Orchid,style={fill=.!40}\\%
Peach,style={fill=.!40}\\%
Periwinkle,style={fill=.!40}\\%
PineGreen,style={fill=.!40}\\%
Plum,style={fill=.!40}\\%
ProcessBlue,style={fill=.!40}\\%
Purple,style={fill=.!40}\\%
RawSienna,style={fill=.!40}\\%
Red,style={fill=.!40}\\%
RedOrange,style={fill=.!40}\\%
RedViolet,style={fill=.!40}\\%
Rhodamine,style={fill=.!40}\\%
RoyalBlue,style={fill=.!40}\\%
RoyalPurple,style={fill=.!40}\\%
RubineRed,style={fill=.!40}\\%
Salmon,style={fill=.!40}\\%
SeaGreen,style={fill=.!40}\\%
Sepia,style={fill=.!40}\\%
SkyBlue,style={fill=.!40}\\%
SpringGreen,style={fill=.!40}\\%
Tan,style={fill=.!40}\\%
TealBlue,style={fill=.!40}\\%
Thistle,style={fill=.!40}\\%
Turquoise,style={fill=.!40}\\%
Violet,style={fill=.!40}\\%
VioletRed,style={fill=.!40}\\%
White,style={fill=.!40}\\%
WildStrawberry,style={fill=.!40}\\%
Yellow,style={fill=.!40}\\%
YellowGreen,style={fill=.!40}\\%
YellowOrange,style={fill=.!40}\\%
}

\usepackage{booktabs}

\newcommand{\eat}[1]{}

\newcommand{\nlstring}[1]{\textit{#1}}

\newcommand{\unkpart}{{\tt UNKNOWN}\xspace}

\newcommand{\dataname}{\textsc{KiloGram}\xspace}

\newcommand{\tokenseq}{\bar{x}}
\newcommand{\token}{x}
\newcommand{\image}{I}
\newcommand{\imageset}{\mathcal{I}}

\newcommand{\clip}{CLIP\xspace}
\newcommand{\vilt}{ViLT\xspace}

\newcommand{\kgfull}{\textsc{Full}\xspace}
\newcommand{\kgdense}{\textsc{Dense}\xspace}
\newcommand{\kgdenseten}{\textsc{Dense10}\xspace}

\newcommand{\expcondition}[1]{\textsc{#1}}
\newcommand{\expcondwhole}{\expcondition{whole}}
\newcommand{\expcondparts}{\expcondition{parts}}
\newcommand{\expcondblack}{\expcondition{black}}
\newcommand{\expcondcolored}{\expcondition{color}}
\newcommand{\expcondaug}{\expcondition{aug}}

\title{Abstract Visual Reasoning with Tangram Shapes}

\author{Anya Ji\textsuperscript{1},~~ Noriyuki Kojima\textsuperscript{1}\footnotemark[1],~~~Noah Rush\textsuperscript{1}\footnotemark[1],~~~Alane Suhr\textsuperscript{1,3}\footnotemark[1], \\ \textbf{Wai Keen Vong\textsuperscript{2},~~ Robert D. Hawkins\textsuperscript{4},} \and \textbf{Yoav Artzi\textsuperscript{1}} \\
  \textsuperscript{1}Cornell University \hspace{8pt} \textsuperscript{2}New York University \hspace{8pt}   \textsuperscript{3}Allen Institute for AI \hspace{8pt} \textsuperscript{4}Princeton  University \\
  \texttt{\{aj592, nk654\}@cornell.edu} \hspace{10pt} \texttt{noahjrush@gmail.com} \\ \texttt{waikeen.vong@nyu.edu} \hspace{10pt} \texttt{suhr@cs.cornell.edu} \\  \texttt{rdhawkins@princeton.edu} \hspace{10pt} \texttt{yoav@cs.cornell.edu}}

\begin{document}
\maketitle
\begin{abstract}
We introduce \dataname, a resource for studying abstract visual reasoning in humans and machines. Drawing on the history of tangram puzzles as stimuli in cognitive science, we build a richly annotated dataset that, with $>$1k distinct stimuli, is orders of magnitude larger and more diverse than prior resources. It is both visually and linguistically richer, moving beyond whole shape descriptions to include segmentation maps and part labels. We use this resource to evaluate the abstract visual reasoning capacities of recent multi-modal models. We observe that pre-trained weights demonstrate limited abstract reasoning, which dramatically improves with fine-tuning. We also observe that explicitly describing parts aids abstract reasoning for both humans and models, especially when jointly encoding the linguistic and visual inputs.

\end{abstract}

\renewcommand{\thefootnote}{\fnsymbol{footnote}}
\footnotetext[1]{Equal contribution, alphabetically ordered.}
\renewcommand*{\thefootnote}{\arabic{footnote}}

\section{Introduction}\label{sec:intro}

\begin{figure}
    \centering
    \includegraphics[width=\linewidth]{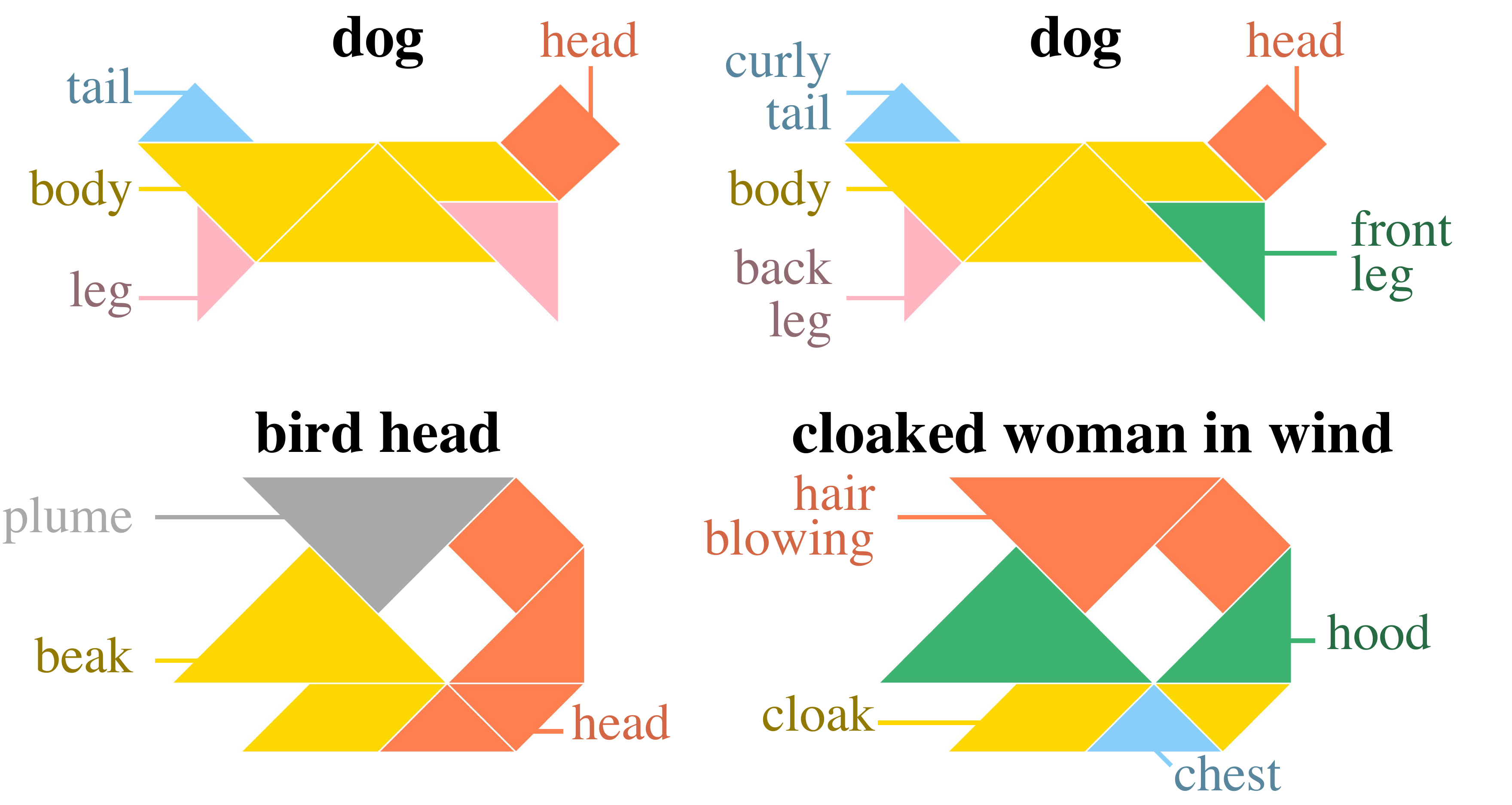}
    \caption{Two example tangrams, each with two different annotations. Each annotation includes a whole-shape description (bold), segmentation to parts (in color), and naming of parts (linked to each part). The top example shows low variability with near-perfect agreement, while the bottom shows high variability with divergence of language and segmentation.}
    \label{fig:intro}
\end{figure}

Reference is a core function of natural language that relies on shared conventions and visual concepts.
For example, in English, a speaker may use the term \nlstring{dog} to refer to a particular animal of the species \emph{canis familiaris}, or, through abstraction, to an object with a less strongly conventionalized name, such as the shape at the top of \autoref{fig:intro}. 
A speaker might refer to such a shape as looking like a \nlstring{dog}, and even point to its parts, like its \nlstring{head} and \nlstring{tail}, despite having few visual features in common with the ordinary referent.

Comprehension and generation of references are critical for systems to engage in natural language interaction, and have been studied extensively with focus on ordinary references~\cite[e.g.,][]{Viethen2008:refexp-gen,Mitchell:10,FitzGerald:13,Mao2015:refexp,Yu:16refmscoco}, in contrast to the visual abstraction illustrated in \autoref{fig:intro}. 
We address this gap by adopting an influential paradigm for probing human coordination in the cognitive science literature: reference games with abstract tangram shapes~\cite[e.g.][]{Clark1986:colab-referring,fox1999listening,Hawkins2020:learning-dynamics-ref-games}.

Unlike photographs of natural objects, where there is often a single canonical label, tangrams are fundamentally ambiguous.
While some shapes fall under strong existing conventions and elicit consensus about appropriate names (e.g.,~\autoref{fig:intro}, top), others are characterized by weaker conventions (e.g.,~\autoref{fig:intro}, bottom) and every speaker may arrive at a distinct but valid description~\cite{zettersten2020finding,hupet1991effects}. 
While such diversity is a key consideration motivating their use as stimuli, existing behavioral studies have typically been limited to a relatively small set of 10--20 shapes, highly restricting the overall diversity of the stimulus class. 
It also limits their applicability for training and analyzing vision and language models, where significantly more data is necessary. 

In this paper, we significantly expand this resource. 
We introduce \dataname,\footnote{\dataname is a portmanteau of \nlstring{kilo} and \nlstring{tangram}.} a large collection of tangrams with rich language annotations.
\dataname dramatically improves on existing resources along two dimensions. 
First, we curate and digitize 1{,}016 shapes, creating a set that is two orders of magnitude larger than collections used in existing work. 
This set dramatically increases coverage over the full range of naming variability, providing a more comprehensive view of human naming behavior. 
Second, rather than treating each tangram as a single whole shape, our images are vector graphics constructed from the original component puzzle pieces. 
This decomposition enables reasoning about both whole shapes and their parts. 

We use this new collection of digitized tangram shapes to collect a large dataset of textual descriptions, reflecting a high diversity of naming behaviors. 
While existing work has focused on naming the complete shape, we also ask participants to segment and name semantically meaningful parts. 
We use crowdsourcing to scale our annotation process, collecting multiple annotations for each shape, thereby representing the distribution of annotations it elicits, rather than a single sample. 
In total, we collect 13{,}404 annotations, each describing a complete object and its segmented parts.

The potential of \dataname is broad. 
For example, it enables the data-driven scaling of studies of human interactions and models of whole-part reasoning in language and vision models.
In this paper, we use \dataname to evaluate the visual reasoning capacities of recent pre-trained multi-modal models, focusing on generalizing concepts to abstract shapes. 
We observe limited generalization of this type in pre-trained models, but significant improvements following fine-tuning with our data. We also see how explicitly referring to and visualizing parts can help reference resolution.  
Data and code, as well as a data viewer are available at: \url{https://lil.nlp.cornell.edu/kilogram/}.

\section{Background and Related Work}\label{sec:bg}

Abstract or ambiguous visual stimuli have been widely used to investigate how human partners coordinate when talking about things in the absence of strong naming conventions going back to \citet{krauss1964changes}. 
Tangrams as stimuli were introduced by \citet{Clark1986:colab-referring}.
These shapes are all built from the same seven primitives, but elicit a wide range of figurative descriptions that conceptualize shapes in different ways~\cite{schober1989understanding,horton2002speakers,duff2006development,holler2011co,horton2012anticipating,ibarra2016flexibility,Shore2018:kthtangrams, atkinson2019social,castillo2019interaction,bangerter2020lexical}. 
It has been observed that some shapes are easier or harder to describe~\cite{hupet1991effects,zettersten2020finding,brashears2020effects}, a property known as \emph{nameability} or \emph{codability}, which has also been studied with non-tangram shapes~\cite[e.g., line drawings;][]{Snodgrass1980:naming-agreement,Cycowicz1997:child-picture-naming,Duabeitia2018:multipic}. 
Even though diversity is a key consideration in working with tangrams, existing stimuli sets are relatively small, limiting their usefulness as NLP benchmarks, where scale is critical. 
Even the largest studies of variability in naming~\cite[e.g.,][]{murfitt2001effect} have used a relatively small set of 60 tangrams. 
\citet{Fasquel2022:tangrams} present a resource that is related and complementary to ours, including 332 PNG-formatted tangrams with whole-shape naming annotations in French.

Contemporary pre-trained vision and language approaches can be categorized along an axis characterizing how they encode the data, from jointly encoding the two inputs~\cite{lu2019vilbert,chen2020uniter,kim:21vilt} to encoding them separately~\cite{radford2021learning,jia2021scaling}. 
Joint encoding aims to capture tighter interaction between the input modalities compared to separate encoding, but is generally more computationally expensive, and can only operate on multi-modal input. 
We study recent models on both ends: \vilt~\cite{kim:21vilt} for joint encoding and \clip~\cite{radford2021learning} for separate encoding.

These models are typically evaluated on image captioning~\cite[e.g.,][]{Chen:15coco} or visual question answering~\cite[e.g.,][]{Antol:15vqa} benchmarks. 
Several benchmarks, such as NLVR~\cite{Suhr:17visual-reason,Suhr2019:nlvr2} and Winoground~\cite{thrush2022winoground}, aim for more focused evaluations with an emphasis on compositionality.
We build on these efforts, but target generalization through abstraction using visually ambiguous stimuli. 
This is inspired by the role of abstraction in human cognition. 
Abstraction is a key step in human perception~\cite{Biederman1987:Recognitionbycomponents} that is critical for generalization~\cite{Gentner1997:StructureMapping,Medin:1993RespectsSimilarity,Shepard1987:lawOfGeneralizatio}, and forms the shared foundation on which human language communication is layered~\cite{Lupyan2018:LanguageIsAbstract,McCarthy2021:LearningCommunicatSharedAbs,Wong2022:IdentifyingConceptLibs}. 
Our focus on part decomposition is aligned with how part identification plays an important role in human abstraction~\cite{Tversky1984:ObjectsPartsCategories}.

\begin{figure}[t]
\centering
\includegraphics[width=\textwidth,clip,trim=10 1 7 10]{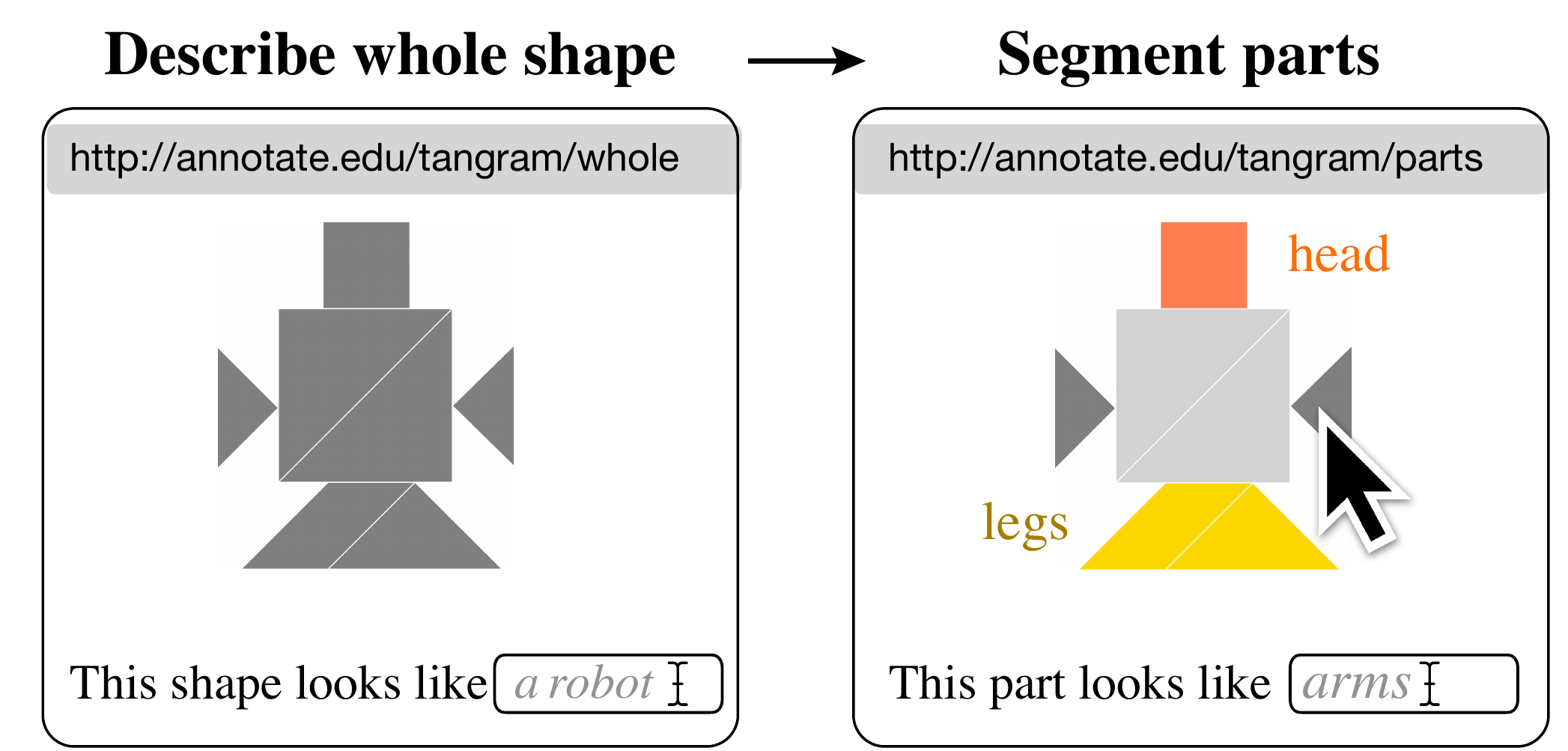}
\caption{The two phases of our annotation task.}
\label{fig:datacollection}
\end{figure}

\section{Data Collection}\label{sec:data}

We scan a large set of tangram puzzles to vector graphics, and crowdsource  annotations of natural language descriptions and part segmentations. 

\subsection{Collecting Tangram Puzzles}\label{sec:data:tangrams}

Tangram puzzles are made of seven primitive shapes~\cite{elffers1999tangram}, which can be combined in a large variety of configurations evoking different concepts. 
We scan 1{,}004 tangrams depicting a broad set of concepts to vector graphic SVGs from \citet{slocum2000tangram}. 
\aautoref{sec:app:moreexamples} shows example tangrams,  \aautoref{sec:app:scanning} details on our process.\footnote{The scanned documents are authorized for use for educational and research purposes (``fair use'') as per U.S. copyright law (Title 17, §108, United States Code). This use does not require permission or usage fees, including for publication. Copyright and use agreement are attached to the data.}
We also manually add 12 tangrams commonly used in previous studies~\cite{Hawkins2020:learning-dynamics-ref-games}.

\subsection{Whole-Part Annotation}\label{sec:data:language}

We design a two-stage crowdsourcing task to elicit natural language English descriptions for each tangram, both of the whole shape and of its parts (\autoref{fig:datacollection}).
First, in the \emph{whole-shape description} stage, the worker is shown a tangram image in grayscale and asked to complete the prompt ``This shape, as a whole, looks like $\_\_\_\_$.''
In the \emph{part annotation} stage, the worker is asked to select one or more puzzle pieces, and complete the prompt ``The part(s) you selected look(s) like $\_\_\_\_$.''
These pieces are then colored and the annotation appears in the corresponding color.
The annotator can delete annotations, annotate a part as \unkpart when they are not sure about its semantics, and add pieces to existing parts.
All pieces must be annotated to submit the task, yielding a complete segmentation map.%

We use Amazon Mechanical Turk for data collection. 
Workers are required to be located in the United States with at least a 98\% HIT acceptance rate, must pass a qualification task, and complete a survey about their language proficiency (see \aautoref{sec:app:crowdsourcing} for further details).
To prevent a small group of workers from dominating the data, each annotator is only allowed to annotate each tangram once, and cannot annotate more than 200 distinct tangrams. 
Workers are paid 0.14 USD per task.\footnote{We set this rate aiming for an hourly rate of 12--15 USD for workers familiar with the task.}

\begin{table}[t]
    \footnotesize
    \centering
    \begin{tabular}{@{}lc@{}}
        \toprule
        \textbf{Mean Description Length} \\
        Whole-shape description & 2.28$\pm$1.62 \\
        Part description & 1.31$\pm$0.77 \\
        \midrule
        \textbf{Vocabulary Size} \\
        Whole-shape description & 3{,}031 \\
        Part description & 3{,}110 \\
        Overall & 4{,}522 \\
        \midrule
        \textbf{Part Segmentation} \\
        Mean parts per shape & 3.63$\pm$1.28 \\ 
        Mean pieces per part & 1.93$\pm$1.20 \\
        \bottomrule
    \end{tabular}
    \caption{Data statistics for the complete dataset.}\label{tab:datastats}
\end{table}

We first collect 10{,}053 annotations for the 1{,}004 scanned tangrams, at least 10 annotations for each tangram (mean=10.01). 
Following this stage of annotation, we collect additional annotations for a subset of the tangrams to create a set with denser language and part segmentation annotation. 
We sample 62 tangrams to be representative of the different levels of diversity in annotations we observe in the initially collected data. 
\aautoref{sec:app:densesampling} describes the sampling procedure.
We also add the 12 tangrams from previous studies for a total of 74 tangrams for dense annotation. 
We conduct additional annotation tasks to have at least 50 annotations for each of the 74 tangrams selected for dense annotation (mean=53.66).\footnote{This includes collecting data from scratch for the 12 commonly used tangrams that were added following the initial collection.} 
The dense annotation gives us a better estimate of the distribution of language for the 74 selected tangrams, for example to use as reference texts in generation tasks. 

In total, we collect 13{,}404 annotations for 1{,}016 tangrams at a total cost of 2{,}172.94 USD.
We lowercase and stem to compute vocabulary size, and tokenize on white spaces to compute description length. \autoref{tab:datastats} shows basic data statistics. 
A total of 297 MTurk workers participate in the annotation, with 98.0\% of the workers speaking English as their first language. 
Those who do not speak English as their first language still rate their English proficiency level as native or close to native. 
1.0\% of the workers speak more than one language, among which the most common are Spanish, German, Japanese, and Chinese.

\subsection{Standard Data Splits}\label{sec:data:split}

We split the dataset for analysis and learning experiments. 
For analysis, we create two overlapping sets: \kgfull  and \kgdense. 
\kgfull includes 1{,}016 tangrams, each with 10--11 annotations (mean=10.11). It includes the 10{,}053 annotations initially collected for the scanned 1{,}004 tangrams. 
For the 12 commonly used tangrams, we sample 10 annotations from the later collection effort. 
\kgdense includes all annotations for the 74 densely annotated tangrams, with at least 50, and 53.66 on average annotations per tangram. 
We also define the set \kgdenseten to include only the annotations from the sparse set for the densely annotated tangrams. 
For learning experiments, we split according to tangrams to create training (692 tangrams), development (125), test (125), and test-dense sets (74).
All densely annotated tangrams are in test-dense. The other three sets are split randomly.

\begin{figure*}[t]
    \centering
    \includegraphics[scale=0.8]{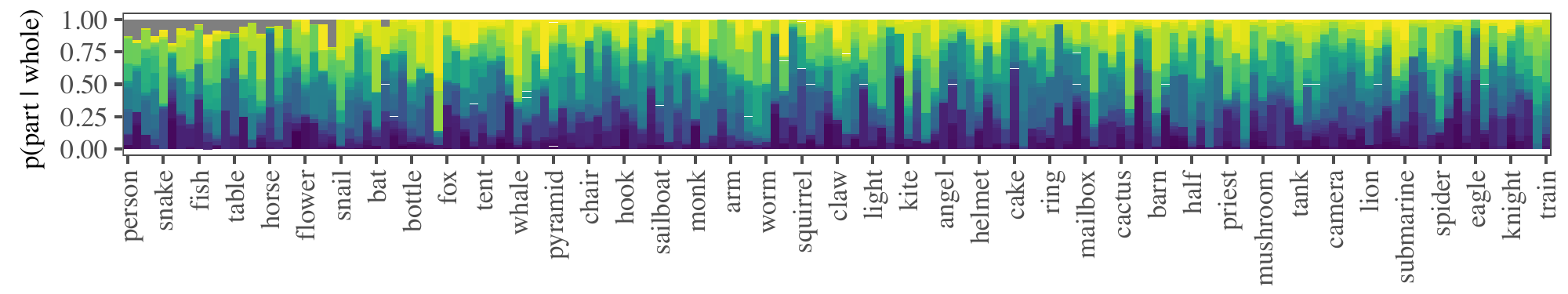}
    \caption{Part distributions for different head words. Whole-shape head words (shown in descending order of frequency from left) elicit a variety of part head word distributions. Colors are randomly assigned to part head words, but are fixed across all bars. Grey indicates part head words with $<0.005$ frequency.}
    \label{fig:word_freq}
\end{figure*}

\begin{table}[t]
    \footnotesize
    \centering
    \begin{tabular}{@{}lccc@{}}
        \toprule
            & \textbf{\kgfull} & \textbf{\kgdense} & \textbf{\kgdenseten} \\ 
            \midrule
         SND & 0.91 $\pm$0.11 & 0.93$\pm$0.06 &0.90$\pm$0.15  \\
         PND & 0.76$\pm$0.19 & 0.79$\pm$0.15 &0.73$\pm$0.20 \\
         PSA &  5.30$\pm$0.62 & 5.09$\pm$0.53 &5.34$\pm$0.77\\
         \bottomrule
    \end{tabular}
    \caption{Mean and standard deviation of our analysis measures on the three sets.}
    \label{tab:analysis}
\end{table}

\begin{figure}[t]
    \centering
    \includegraphics[width=\linewidth]{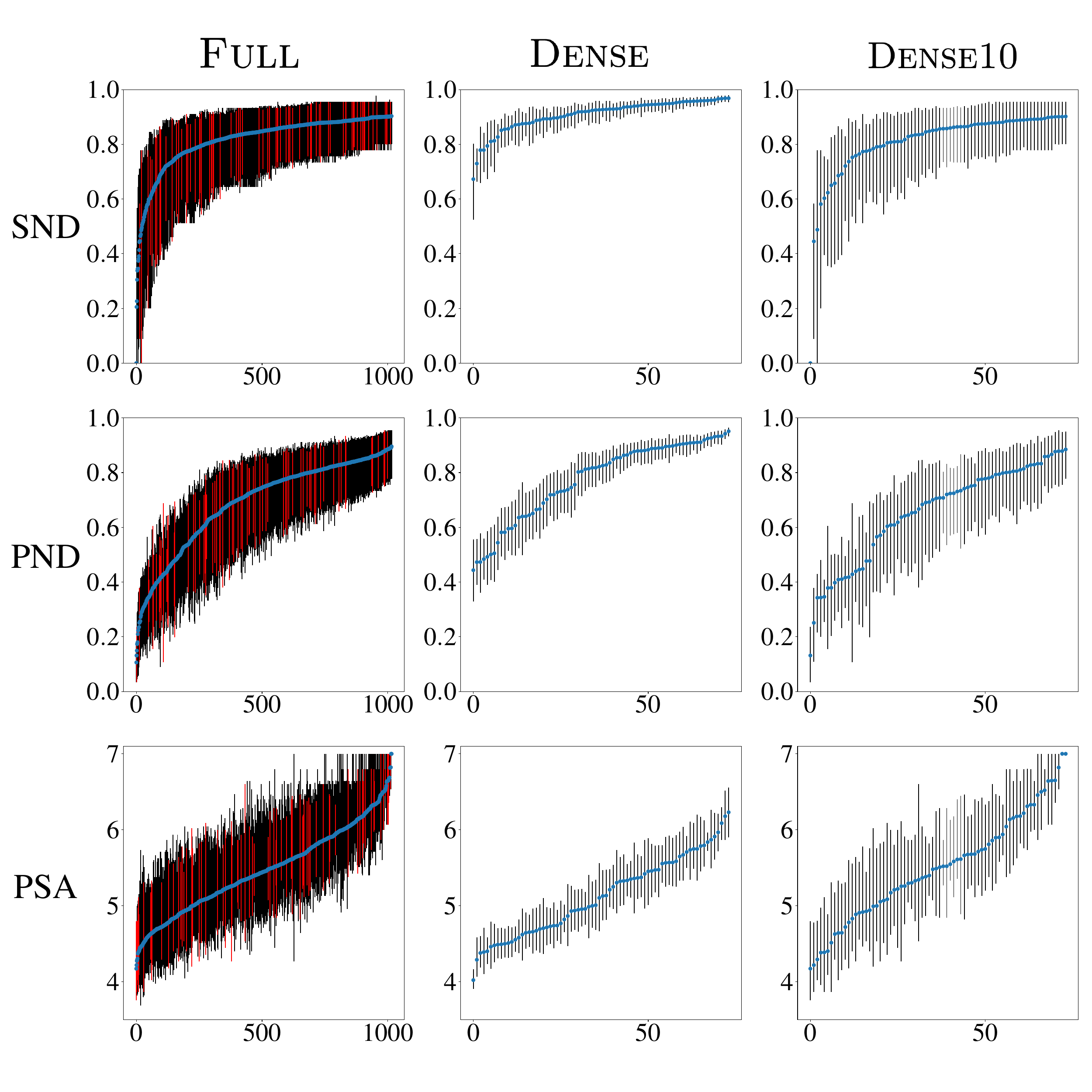}
    \caption{Per tangram SND, PND, and PSA mean values and 95\% confidence interval. Tangrams are ordered along the $x$-axis in ascending order according to the plotted measure. Values are calculated by bootstrapping with 1{,}000 resamplings. In the \kgfull plots, the 74 densely annotated tangrams are colored red.}
    \label{fig:measures}
\end{figure}

\section{Data Analysis}\label{sec:analysis}

The language and concepts annotators use reflect varying degrees of consensus around conventions for describing the appearance of shapes and their parts. 
For analysis, we preprocess the annotations by lowercasing, tokenizing, lemmatizing, and removing stop words using NLTK~\cite{Bird2004:nltk}. 
We use the larger \kgfull set for our analyses (\autoref{sec:data:split}), unless otherwise noted. 

For a broad overview of the types of concepts evoked, we manually tag 250 randomly sampled annotations: 30.8\% use human-like concepts (e.g., \nlstring{dancer}), 31.2\% animate but non-human concepts (e.g., \nlstring{dog}), and 38.0\% non-animate concepts (e.g., \nlstring{house}). 
We examine how part words differ across whole-shape concepts by extracting head words from whole-shape and part descriptions. \autoref{fig:word_freq} shows the distribution of part head words for each of 272 whole-shape head words with $>$10 occurrences, ranked in order of frequency. %
\autoref{fig:tangrams:headpart} in the appendix illustrates how the most common part word \nlstring{head} is used in different tangrams.

A central problem of visual abstraction is the degree of ambiguity or subjectivity that a  shape evokes across different people~\cite{murthy2022shades}: some descriptions have higher consensus than others.
We define three measures of variability along different dimensions: shape naming divergence (SND), part naming divergence (PND), and part segmentation agreement (PSA).
\autoref{tab:analysis} lists the mean and standard deviation for these three measures over the sparsely and densely annotated data.

\paragraph{Shape Naming Divergence (SND)}
A tangram's SND quantifies the variability among whole-shape annotations. 
SND is an operationalization of nameability, a criteria that is commonly used to measure how consistent is naming of an object across individuals~\cite[e.g.,][]{zettersten2020finding}. 

Formally, a whole-shape annotation is a sequence of $M$ tokens $\tokenseq = \langle \token_1,\dots,\token_M\rangle$. 
Given a tangram with $N$ annotations $\tokenseq^{(j)}$, $j=1,\dots,N$, each of length $M^{(j)}$, we define $w^{(j)}_i$ for each token $\token_i^{(j)}$ in annotation $\tokenseq^{(j)}$ as the proportion of other annotations of that tangram that do not contain $\token_i^{(j)}$:
\begin{equation}
    w^{(j)}_i = \frac{1}{N-1}\sum_{j'=1}^N \mathbbm{1}[\token^{(j)}_i \notin \tokenseq^{j'}]\;\;,
\end{equation}
where $\mathbbm{1}$ is an indicator function. 
The divergence of annotation $\tokenseq^{(j)}$ is $W^{(j)} = \frac{1}{M^{(j)}}\sum_{j=0}^k w^{(j)}_i$. 
The divergence of a tangram is $W = \frac{1}{N}\sum_{j=0}^N W^{(j)}$. 
For example, the SNDs of the tangrams in \autoref{fig:intro} computed only with the two annotations displayed are 0.00 (top) and 1.00 (bottom).

\begin{figure*}
\centering
\includegraphics[width=0.9\textwidth]{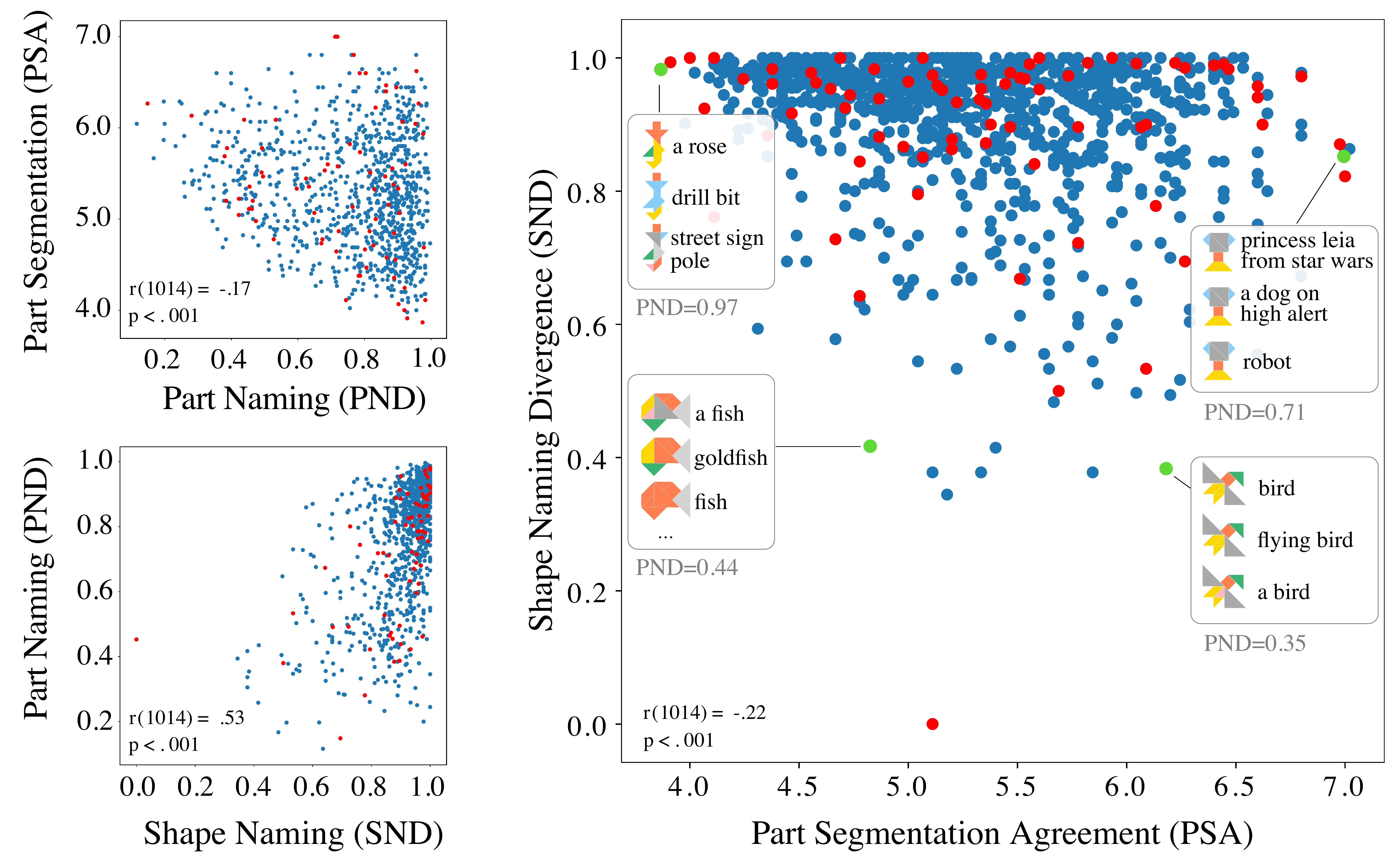}
\caption{SND, PND, and PSA correlations computed over the \kgfull set. Representative examples of different SND and PSA values are illustrated on the right. Densely annotated examples are highlighted in red.}
\label{fig:correlations}
\end{figure*}

Mean SND is relatively high in our data, with 0.91 on \kgfull (\autoref{tab:analysis}). 
We observe relatively similar values for \kgdense and \kgdenseten, albeit with lower standard deviation for \kgdense, as expected with more annotations. 
Annotators often use words that are unique to their annotation. 
We observe perfect consensus for only one tangram, and mostly similar annotations with relatively few deviations for a few others. \autoref{fig:correlations} shows several examples.

\paragraph{Part Naming Divergence (PND)}
SND measures annotation divergence for part name annotations collected in the second step of the annotation task. PND is computed identically to SND, but with the concatenation of all part names of an annotation as the input text $\tokenseq$. 
For example, the PNDs of the two tangrams in \autoref{fig:intro} computed with only the two annotations displayed are 0.19 (top) and 1.00 (bottom). 
In general, part descriptions are more similar than whole-shape descriptions with mean PND of 0.76 (\autoref{tab:analysis}).

\paragraph{Part Segmentation Agreement (PSA)}
Annotators segment the tangrams into parts by grouping the tangram puzzle pieces. 
PSA quantifies the agreement between part segmentations as the maximum number of pieces that does not need to be moved to another group in order to edit one segmentation to another. 
We compute PSA as a linear sum assignment problem with maximum weight matching. 
For each pair of segmentations, we create a cost matrix, where the number of rows is the number of parts in one annotation and the number of columns is the number of parts in the second annotation. 
The value of each matrix element is the number of matching puzzle pieces between the two corresponding parts in the two annotations. 
The tangram PSA is the mean of costs for all annotation pairs. 
For example, the PSAs of the two tangrams in \autoref{fig:intro} computed with only the two annotations displayed are 6.00 (top) and 3.00 (bottom).

The mean PSA in our data is 5.30 (\autoref{tab:analysis}), with an approximately normal distribution of values. 
Some tangrams have strong segmentation cues, such that annotators reach perfect consensus, while others  elicit significant segmentation disagreement.

\paragraph{Dense Annotations}

The comparison of \kgfull, \kgdense, and \kgdenseten illustrates how well our data approximates the real distribution of annotations for each tangram, and the advantage of \kgdense.  
\autoref{fig:measures} shows the complete distribution of values. 
Comparing \kgdenseten and \kgdense, the rankings of the tangrams are largely the same with the additional annotations: for SND, Spearman's rank correlation coefficient is $r(72)=.78$, $p\ll.001$; for PND, $r(72)=.87$, $p\ll.001$; for PSA, $r(72)=.76$, $p\ll.001$. The tangrams sampled for \kgdense represent well the distribution of tangrams along the different measures, as illustrated by the red highlights in \autoref{fig:measures}.

\paragraph{Inter-measure Correlations}

\autoref{fig:correlations} illustrates the correlations between the three measures. 
The divergences of the two types of language annotations, whole-shape and part descriptions, show moderate positive correlation $r(1014)=.531$, $p\ll.001$. 
This indicates that tangrams that are annotated with similar whole-shape descriptions are often annotated with similar part descriptions. 
Nevertheless, many tangrams with similar whole-shape descriptions have diverse part descriptions. 
The correlations between language annotation divergence and PSA are lower, $r(1014)=-.216$, $p\ll.001$ for SND and PSA and $r(1014)=-.165$, $p\ll.001$ for PND and PSA.

\section{Visual Reasoning with Tangrams}\label{sec:absvis}

We use \dataname to evaluate the reasoning of \clip~\cite{radford2021learning} and \vilt~\cite{kim:21vilt} through a reference game task, where the model is given a textual description and selects the corresponding image from a set of images.
Formally, given a textual description $\tokenseq$ and a set of $k$ images $\mathcal{I} = \{\image_1,\dots,\image_k\}$, the task is to select the image $\image_i \in \mathcal{I}$ corresponding to $\tokenseq$. 
We cast the task as computing a similarity score $f(\tokenseq, \image_i)$ between the description $\tokenseq$ and an image $\image_i$. We select the corresponding image as $\image^* = \arg\max_{\image_i \in \imageset}f(\tokenseq, \image_i)$. 

\subsection{Reference Game Generation}\label{sec:absvis:gamegen}

We randomly generate reference games for an annotated text-image pair $(\tokenseq, \image)$ by sampling additional $k-1$ images from data under several constraints. 
We do not include repeating images in the set of $k$ images or images that have identical whole-shape text annotations. 
This avoids obvious ambiguity that is impossible to resolve in the target selection. 
We also require all images to be annotated with the same number of parts. This reduces the chance of the model relying on simple part counting to discriminate between target images when including parts in the text (condition \expcondparts{} below). 
\aautoref{sec:app:randomrefgames} shows the impact of these constraints through analyzing experiments not using them.

\begin{figure*}
    \centering
    \includegraphics[width=0.95\textwidth,clip,trim=50 800 50 20]{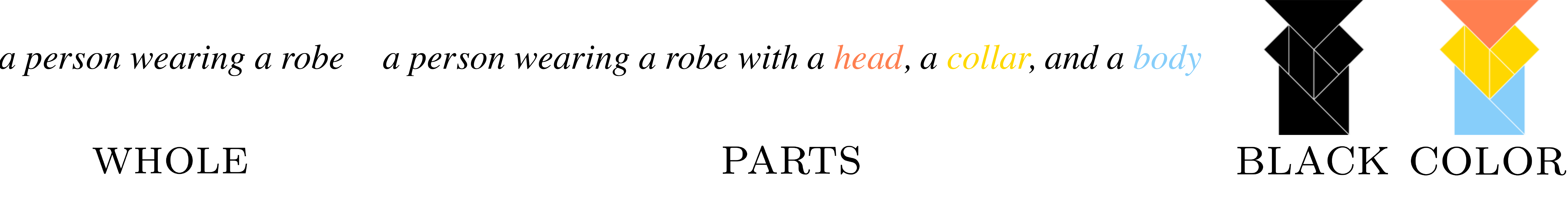}
    \caption{Illustration of the language and vision modalities under the different experimental conditions.}\label{fig:inputshort}
\end{figure*}

\subsection{Models}\label{sec:reasoning:models}

We instantiate $f$ using \clip or \vilt, two models based on the Transformer architecture~\cite{vaswani2017attention}.  
We provide a brief review of the models, and refer the reader to the respective papers for further details. 

\clip uses two separate encoders to generate separate fixed-dimension representations of the text and images. 
It uses contrastive pre-training with a symmetric cross entropy loss on a large amount of aligned, but noisy web image-text data. 
We implement the scoring function $f$ with \clip by encoding the text $\tokenseq$ and all images $\image \in \imageset$ separately, and then computing the dot-product similarity score of the text with each image. This is identical to the \clip pre-training objective, which potentially makes \clip suitable for our task out of the box. 

\vilt uses a single encoder that takes as input both the text and image inputs together. 
\vilt pre-training also uses aligned image-text data, but from existing benchmarks~\cite{Lin:14mscoco,Krishna:16,Ordonez:11sbu,Sharma:18gcc}. 
It is pre-trained using multiple self-supervised objectives, including image-text matching via a binary classification head, which is suitable for our task out of the box. 
We implement $f$ using this classification head. Given a text $\tokenseq$ and an image $\image \in \imageset$, we compute their similarity using the matching classification head. 

\subsection{Experimental Conditions}\label{sec:absvis:expcond}

We study several input variants. 
\autoref{fig:inputshort} illustrates the modalities under the different conditions, and \aautoref{sec:app:condexamples} shows complete example inputs.  
For the textual description $\tokenseq$, we experiment with including the whole-shape description only (\expcondwhole) or adding part names (\expcondparts) by combining with the whole-shape description using the template \nlstring{<whole shape> with <part>, <part>, ..., and <part>}. This tests the ability of models to benefit from part names. 
We consider two image $\image$ conditions: coloring all parts with the same color (\expcondblack) or coloring parts differently (\expcondcolored). The color choice in \expcondcolored{} corresponds to the position of the part name in $\tokenseq$, when the text includes part names (\expcondparts). 

We experiment with the original pre-trained model weights, and with contrastive fine-tuning on our data using a symmetric cross entropy loss~\cite{radford2021learning}. 
During fine-tuning only, we consider a data augmentation condition (\expcondaug), where we augment the data by creating examples that include only a subset of the part names in the text and coloring only the parts corresponding to the included parts names in the image, while all other parts remain black. We generate partial part examples for all possible subsets of parts for each example. \aautoref{sec:app:condexamples} illustrates the generated examples. 
When generating reference games for the augmented data, we constrain all the examples  within a reference game to have the same number of parts in their full annotation, otherwise the task could be solved by counting parts.
Part names are shuffled when creating the augmented data, and part colors correspond to the sequential position of the part name in the templated text.

\subsection{Implementation Details}\label{sec:absvis:impl}

We set the size of the reference game context to $k=10$ throughout our experiments. During contrastive fine-tuning, we create a text-image matching matrix of size $k\times k$ for each generated reference game in our training data by  randomly selecting a text description for each tangram distractor from its annotations. We compute matching loss  in both directions, from text to images and vice versa.
In practice, this is equivalent to creating $2k$ reference games in both directions, and provides more informative updates. 
For all experiments, we use an ensemble of three models combined by element-wise multiplication of their outputs.  
\aautoref{sec:app:implmodels} provides model-specific implementation details. 
\aautoref{sec:app:reproduce} provides a reproducibility list.

\begin{figure*}[t]
    \begin{floatrow}
    \capbtabbox{%
        \footnotesize\centering
        \setlength{\tabcolsep}{5pt}
        \begin{tabular}{@{}lcccccccc@{}}
            \toprule
            \multirow{2}{*}{\textbf{Condition}} && \multicolumn{2}{c}{\clip}  && \multicolumn{2}{c}{\vilt} && Human \\
            \cmidrule{3-4} \cmidrule{6-7} 
            && PT & FT && PT & FT \\
            \midrule
            \multicolumn{7}{@{}l}{\textbf{Development Results}} \\
            \midrule
            \expcondwhole+\expcondblack && 16.1 & 43.3 && 12.9 & 40.9 && 47.7 \\
            \expcondparts+\expcondblack && 16.4 & 45.3 && 12.5 & 45.7 && 49.1\\
            \expcondwhole+\expcondcolored && 15.9 & 40.8 && 11.7 & 41.0 && 49.5\\
            \expcondparts+\expcondcolored && 15.0 & 45.4 && 10.7 & 75.2 && 63.0 \\
            \expcondparts+\expcondcolored+\expcondaug  && -- & 47.6 && -- & 72.2 \\
            \midrule
            \multicolumn{7}{@{}l}{\textbf{Held-out Test Results}} \\
            \midrule
            \expcondwhole+\expcondblack && 17.9 & 42.5 && 13.1 & 44.5 \\
            \expcondparts+\expcondblack && 18.6 & 45.8 && 13.3 & 50.3 \\
            \expcondwhole+\expcondcolored && 18.1 & 41.4 && 12.8 & 44.8 \\
            \expcondparts+\expcondcolored && 17.0 & 46.5 && 11.7 & 77.3  \\
            \expcondparts+\expcondcolored+\expcondaug  && -- & 50.2 && -- & 74.4 \\
            \bottomrule
        \end{tabular}%
    }{%
        \caption{Reference game accuracies (\%) for the different experimental conditions with pre-trained (PT) or fine-tuned (FT) models, as well as for human subjects.}\label{tab:performance}%
    }%
    \ffigbox[\FBwidth]{%
        \centering%
        \includegraphics[width=0.86\linewidth]{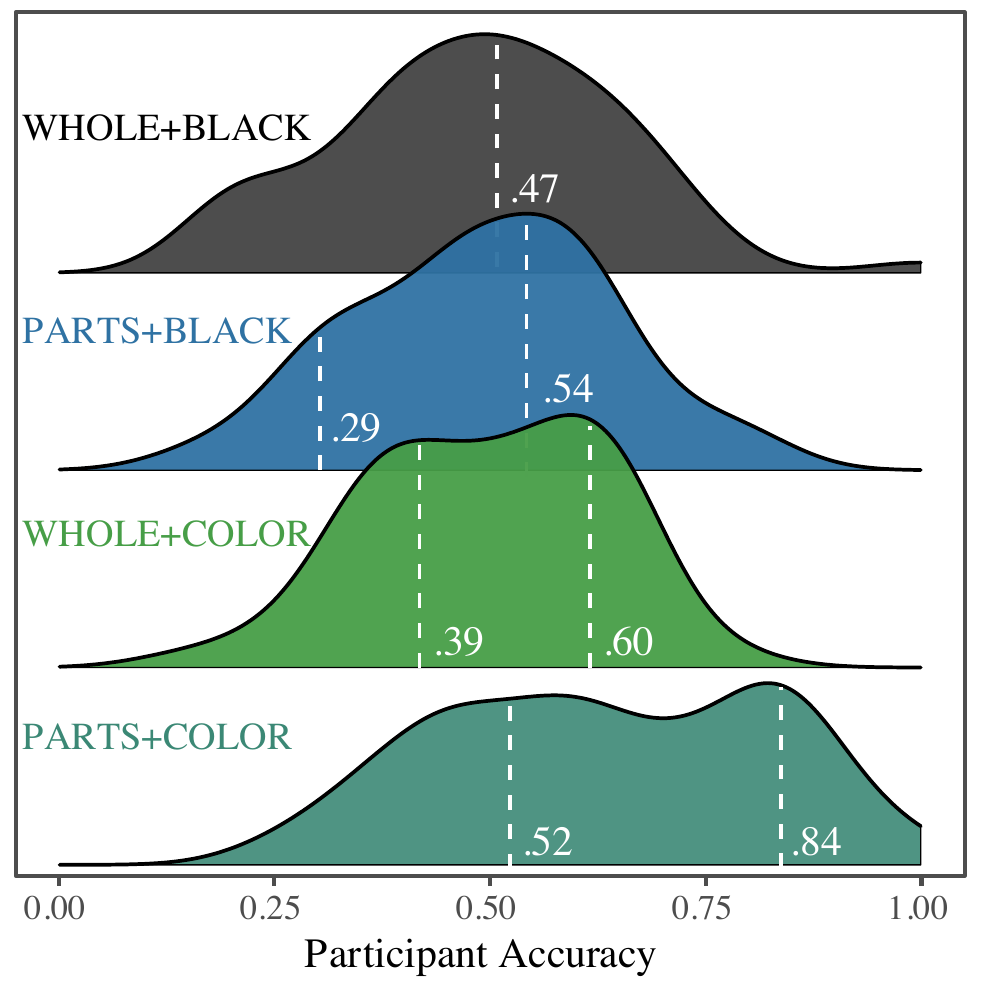}%
    }{%
    \caption{The distribution of each human participant's mean accuracy in the four conditions. The white dashed lines are the estimated means of a two-component Gaussian mixture model.}\label{fig:humandist}%
    }
    \end{floatrow}
\end{figure*}

\subsection{Estimating Human Performance}\label{sec:exp:humans}

We conduct an initial estimation of expected human performance on the same evaluation task by recruiting an independent group of 217 human participants. 
Each participant is randomly assigned to one of the four conditions and shown a random sequence of 20 trials from that condition, preventing leakage across conditions.
On each trial, we present an annotation from our development set along with the corresponding context of ten tangrams and ask the participant to click the tangram that was being described. 
We randomly sample one referential context per annotation, which provides coverage over all 125 tangrams and over 600 unique descriptions in each condition. Before the actual test trials, each participant is provided with a fixed set of 10 practice trials with feedback indicating whether they have selected the correct tangram, and if not, we highlight the correct answer. 
Performance in the practice trials is not considered in our analysis. 
\aautoref{sec:app:human} provides further details.

\subsection{Results and Analysis}\label{sec:results}

\Cref{tab:performance} shows development and test reference game accuracies under different experimental setups, including for human studies. \autoref{fig:humandist} shows the accuracy distribution for human participants. 

While both models perform better than a random baseline (10\%) out of the box, we generally observe poor performance with the pre-trained weights (PT).
\clip slightly outperforms \vilt  throughout, potentially because it is trained with a contrastive objective similar to a reference game. Whereas \vilt's matching loss is aligned with our goal, it is only one of several losses in its objective. 
We observe no reliable improvement from adding part information, either textual or visual. 
The low performance on \expcondwhole+\expcondblack{} indicates the models fail to generalize familiar concepts to abstract shapes and the lack of consistent improvement with part information indicates an inability to reason about the correspondence of text and colored parts. 

Fine-tuning (FT) dramatically improves performance for both models. 
Adding part names to the text description improves both models (\expcondparts+\expcondblack). 
However, segmentation information in the form of part coloring without part names (\expcondwhole+\expcondcolored) shows no benefit. 
Although \vilt does not benefit from color information alone, the combination with part names (\expcondparts+\expcondcolored) shows significant added improvement in performance over having access to part information in one of the modalities. 
Overall, we observe small consistent differences in performance between the two models, except when having access to both part names and colors (\expcondparts+\expcondcolored), which \vilt effectively uses following fine tuning. 
This may be because \vilt's tight integration of the modalities in its single encoder allows it to take advantage of the part correspondence information provided when both part names and colors are given. 

Human performance follows a similar trend to the fine-tuned models: adding part names and segmentation helps performance, and their benefit is most pronounced when both are provided. 
Human performance is significantly higher than pre-trained (PT) models across all four conditions. 
Fine-tuning (FT) closes this gap.
Indeed, in the \expcondparts+\expcondcolored{} condition, \vilt significantly outperforms mean human performance. 
To better analyze human results, we fit a two-component Gaussian mixture model to the distribution of individual participants' accuracies (\autoref{fig:humandist}). 
We observe two components for all conditions except \expcondwhole+\expcondblack, indicating two distinct sub-populations.
For example, for \expcondparts+\expcondcolored, the low-performing sub-population has a mean accuracy of 52.5\%, while the high-performing has a mean of 83.8\%, significantly outperforming the fine-tuned \vilt. 
It is possible that the lower-performance sub-population is not making full use of the additional information.

\begin{figure*}[t]
    \centering
    \footnotesize
    \includegraphics[width=0.45\linewidth]{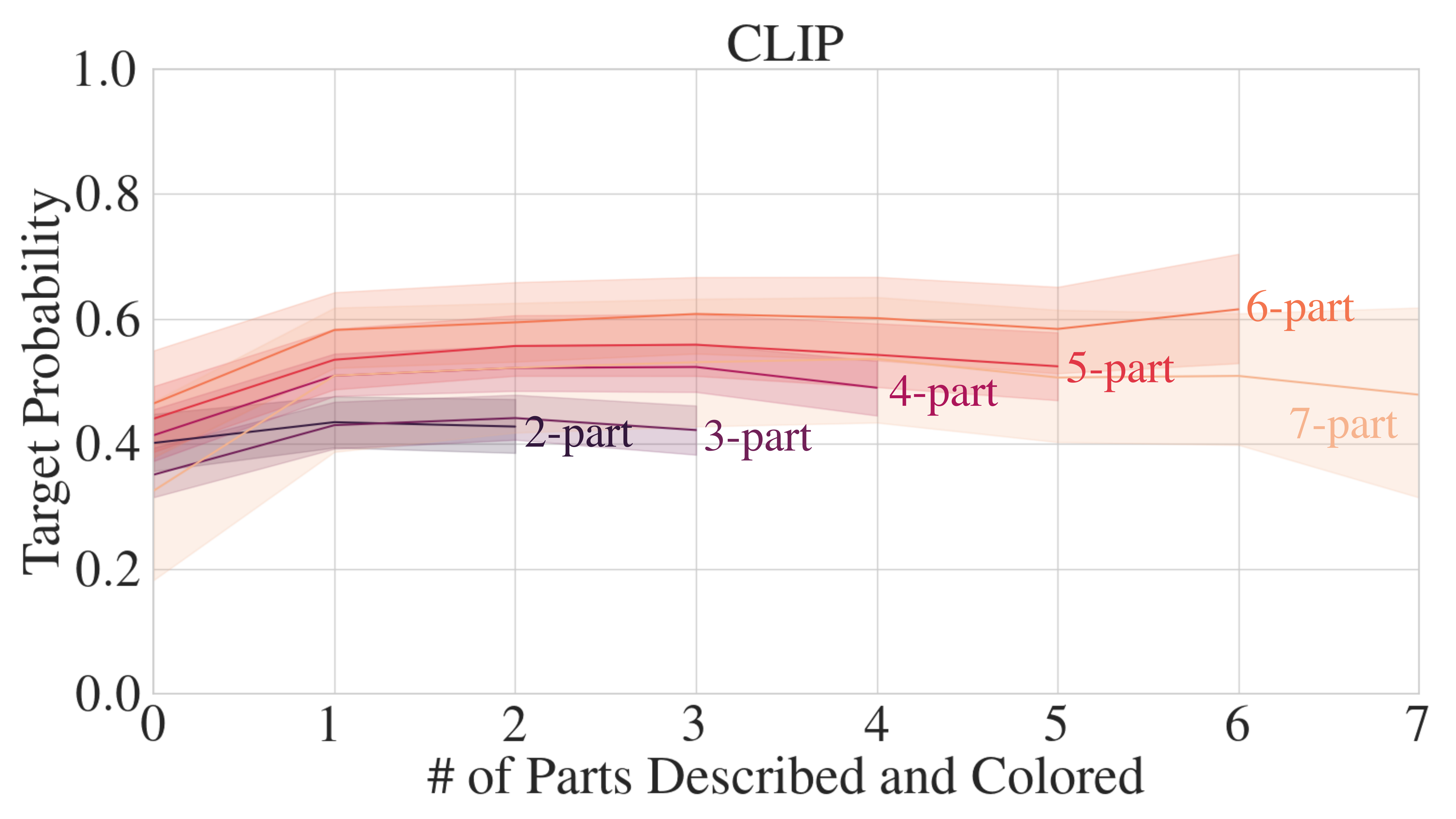}
    \medskip
    \includegraphics[width=0.45\linewidth]{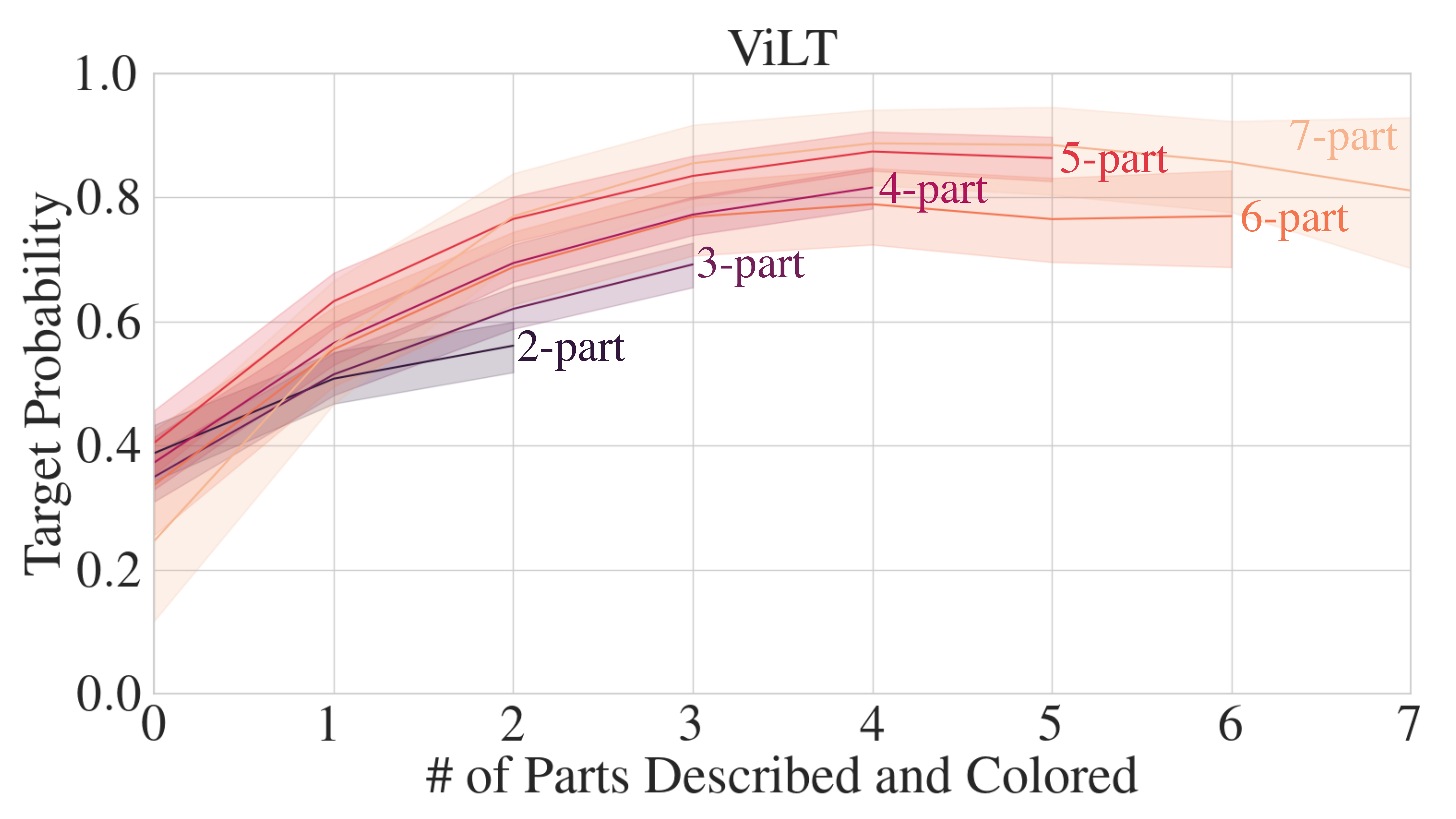}
    \caption{Mean probability assigned to the correct image using fine-tuned \clip (left) or fine-tuned \vilt (right) on the development set, by number of parts included in text and colored in the images. Curves are separated by total number of parts in the annotation of the target example. Error bands are bootstrapped 95\% confidence intervals.}\label{fig:partplots}
\end{figure*}

Data augmentation (\expcondaug) improves performance for \clip, but not for \vilt, which even shows a small decrease in performance, although still significantly outperforming \clip. 
We hypothesize that the presence of training examples with partial part information complicates resolving the correspondence between parts and their name, resulting in overall lower \vilt performance. 
We leave further study of this hypothesis for future work. 

The augmentation condition fine-tunes the models to handle examples with partial part information, and allows to study the impact of gradually adding part information. 
We apply the augmentation process to the development data to generate the data for this analysis. 
\autoref{fig:partplots} shows the effect of gradually adding part information on the probability of the correct prediction, separated by the total number of parts in the example. 
Overall, part information is beneficial, but with a diminishing return as more part information is added. 
We observe this for both models, but with a much faster rate for \clip, which overall shows much lower performance. 
\vilt is able to benefit from increasing part information, with the benefit diminishing only after four parts are provided.

\section{Discussion}\label{sec:disc}

\dataname provides a new window into the visual abstraction capacity of grounded language models and their ability to generalize concepts beyond their photographic appearance, an integral component of human concept representations~\cite{Fan2015:CommonObjectRep}. 
Our experiments show that there is significant room to improve pre-trained models,  which should be able to perform zero-shot reference game tasks without fine-tuning as well as humans do~\cite{Clark1986:colab-referring}. 
The improved performance after fine-tuning indicates the multi-modal architecture itself has the potential for higher performance, which current pre-training regimes likely do not support. 
In particular, \vilt's improved performance as a function of additional part information suggests that more structured concept alignment may play a role in this effort (e.g., between parts expressed as lexical items and the corresponding elements of the image).

While we focused on the task of reference resolution, \dataname is also well-suited for production tasks (e.g., generating human-like distributions of descriptions or coloring named parts on a blank tangram) as well as instruction-following tasks (e.g., placing pieces in the described configuration to reconstruct a tangram). 
More broadly, our data emphasizes the need for maintaining well-calibrated distributions over the many different possible ways that people may conceptualize or talk about things, rather than collapsing to a ``best'' prediction.

\section{Limitations}\label{sec:limit}

Although randomly constructed reference games provide an interpretable evaluation metric, they also pose several limitations.
Performance is limited by the fact that descriptions were elicited for isolated images.
These descriptions do not reflect the kind of pragmatic reasoning commonly deployed by human speakers in reference games to resolve ambiguities~\cite{Goodman2016:PragmaticProbInf}.
In other words, annotators were not able to anticipate the necessary level of detail to disambiguate the object from a specific context of distractors, hence the descriptions may be underinformative.
Randomly generated reference games may include ambiguities that make them impossible to solve (e.g., two objects that could both plausibly be described as a \nlstring{bird}).
The possible performance ceiling on these games is likely below 100\%.
Extending the data through interactive reference games is an important direction for future work. 
Likewise, our studies of baseline human performance on this task are preliminary. 
We found that participants clustered into higher- and lower-performing groups, likely reflecting attentional and motivational factors (e.g., some participants may not have fully attended to the provided part information). 
A better understanding of human behavior is critical before making any clear conclusions comparing humans and model performance. 
Ultimately, models only outperformed mean human performance significantly only after fine-tuning on approximately 6{,}600 example reference games.

Our resource contribution and analysis are focused on English. 
While the data collection design does not make language-specific assumptions, it depends on the availability of proficient speakers, which is limited in contemporary crowdsourcing services for certain languages. 
Our large collection of visual stimuli is well suited to extend our data collection to other languages and cultures, which may display different abstractions. This is an important direction for future work. 
Extending our analysis to other languages depends on the availability of pre-trained models in these languages, which may be limited by the availability of aligned language vision data and the computational resources required for pre-training.

\section*{Acknowledgements}

This research was supported by ARO W911NF21-1-0106, NSF under grant No. 1750499, and a gift from Open Philanthropy. 
NK is supported by Masason Fellowship, AS by a Facebook PhD Fellowship and an NSF GRF under grant No. 1650441, and RDH by a CV Starr Fellowship.
We thank Rob Goldstone, Judith Fan, Cathy Wong, and the anonymous reviewers for their helpful comments and suggestions. 
We are grateful for the contributions of the workers on Mechanical Turk.

\bibliography{main,local}
\bibliographystyle{acl_natbib}

\clearpage

\appendix

\section{Appendix}\label{sec:app}

\renewcommand{\thefigure}{A.\arabic{figure}}
\renewcommand{\thetable}{A.\arabic{table}}
\setcounter{table}{0}
\setcounter{figure}{0}

\subsection{Examples from \dataname}\label{sec:app:moreexamples}

\begin{figure*}[t]
    \centering
    \includegraphics[scale=0.6]{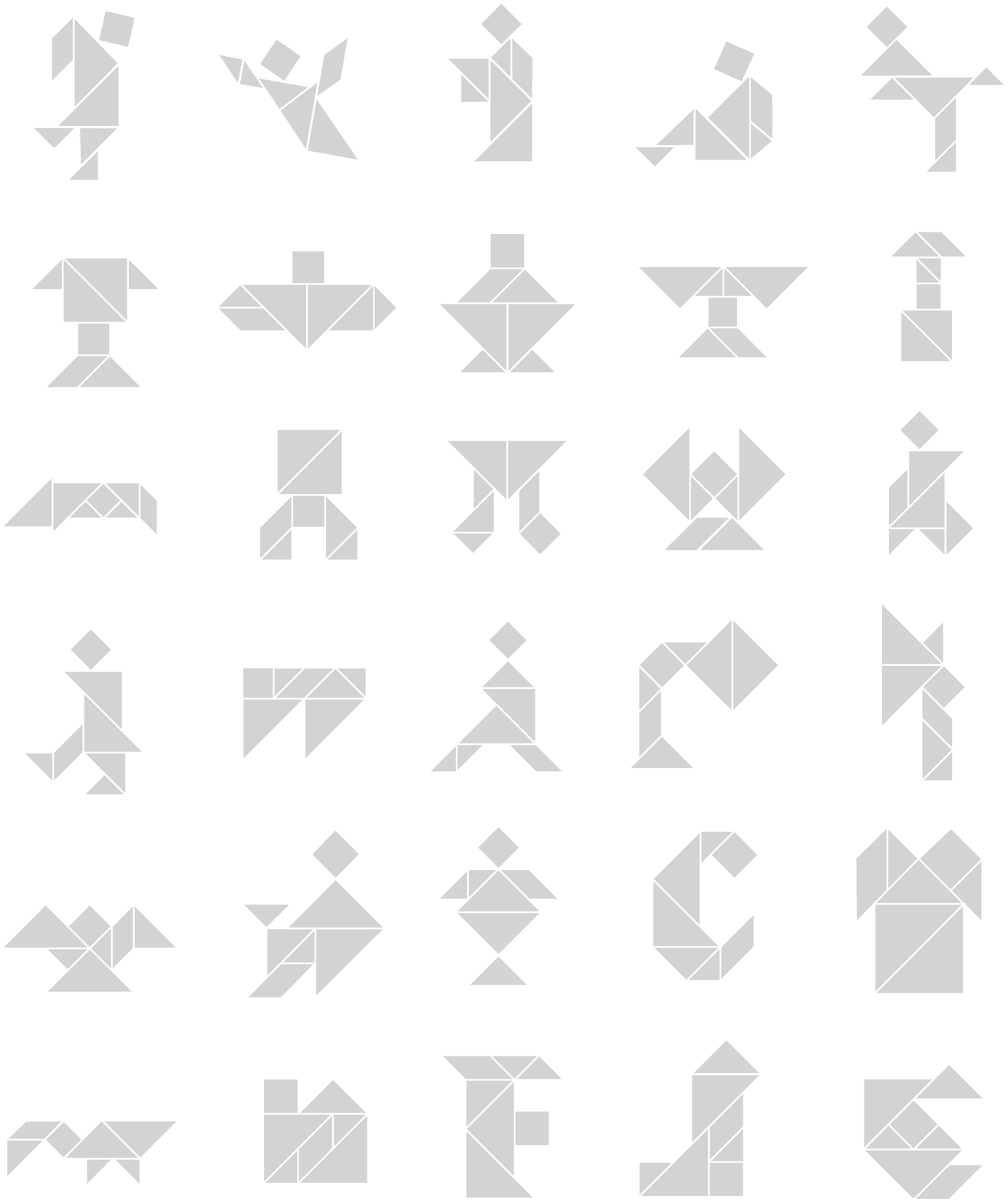}
    \caption{Example tangrams from our dataset.}
    \label{fig:tangramexamples}
\end{figure*}

\begin{figure*}[t]
    \centering
    \includegraphics[scale=0.6]{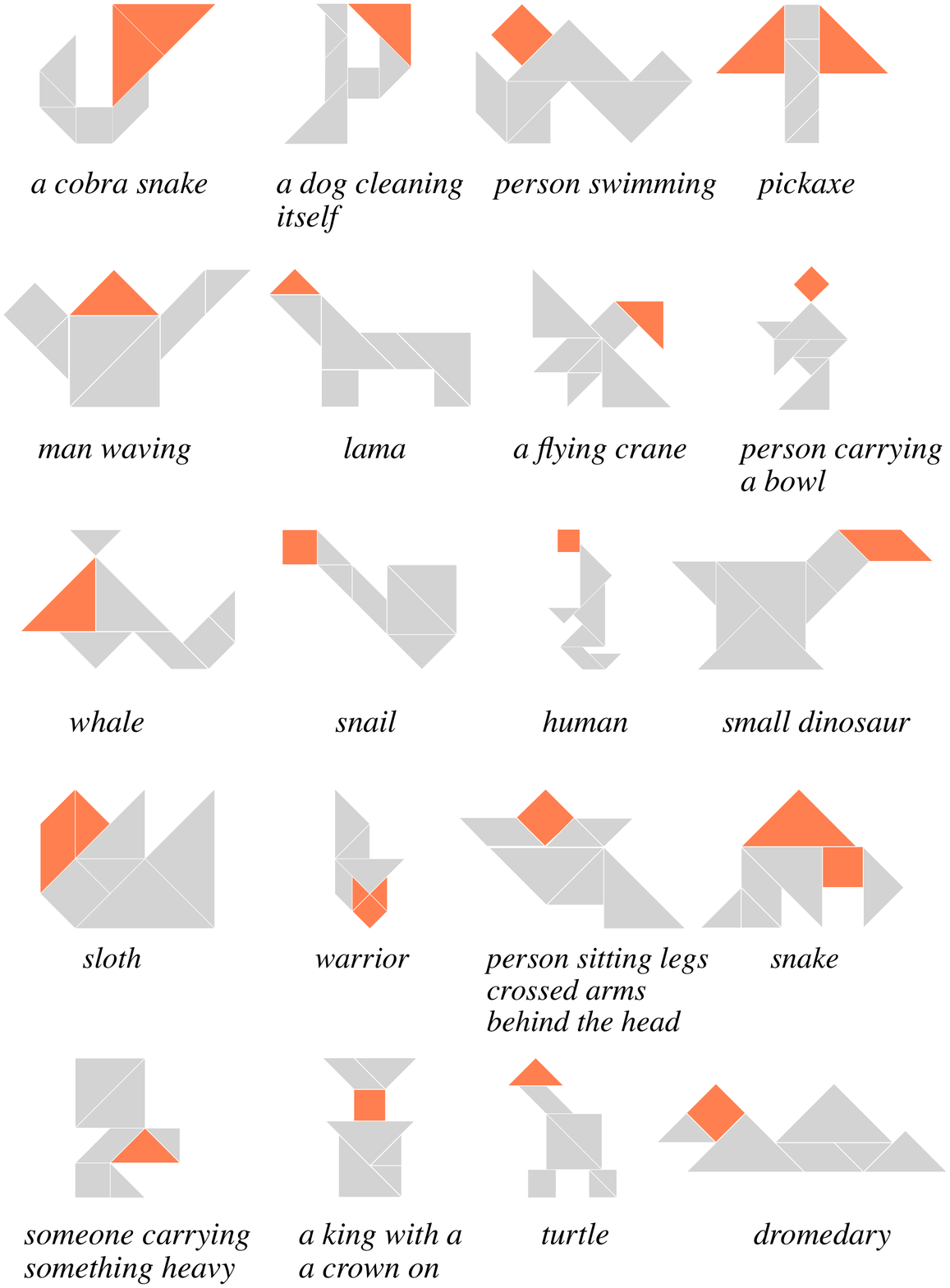}
    \caption{Example tangrams containing the part description \nlstring{head}. Each example includes a tangram and its whole-shape description. We highlight the segmentation corresponding to \nlstring{head} in each tangram.} 
    \label{fig:tangrams:headpart}
\end{figure*}

\autoref{fig:tangramexamples} shows example tangrams from our data.
\autoref{fig:tangrams:headpart} shows examples of the use of the part name \nlstring{head}, the most common part head word in the data. 
All data can be browsed on the data visualization dashboard: \url{https://lil.nlp.cornell.edu/kilogram/}.

\subsection{Collecting Tangrams}\label{sec:app:scanning}

We scan all the pages of tangram solutions from \citet{slocum2000tangram} into JPEG files to extract SVG files of individual tangrams. 
We use heuristics based on edge and corner detection~\cite{harris1988combined} to extract individual tangrams into separate files by detecting the four corners of each puzzle and adding padding.\footnote{We use OpenCV for this process~\cite{opencv_library}.}
We heuristically detect the individual standard pieces in each tangram using corner detection. Because the shapes are standard, we can test if an extracted shape is an expected puzzle's piece and if we obtain the expected number of such shapes. 
We resize each tangram and all its pieces to a standard size, and label the ID of each puzzle piece consistently across all tangrams. We heuristically and manually validate the outputs, and prune solutions that fail to vectorize properly, for example if the process fails to recover exactly seven pieces.

\subsection{Crowdsourcing Qualifications and Survey}\label{sec:app:crowdsourcing}

The qualifier includes three multiple choice questions aimed to ensure that (a) the annotator describes the abstract shape meaningfully instead of simply describing its geometry; (b) each part description only contains one part (\nlstring{body} and \nlstring{arms} instead of \nlstring{body with arms}); and (c) the part descriptions correspond to the description of the whole shape. 
We provide a short video tutorial of the task and examples of invalid annotations for workers to view before completing the qualifier. We also collect basic non-identifying demographic data from each worker, including the languages that they speak and their proficiency, if English is their first language, and where they learned English.
We retain the correspondence of anonymized hashed worker IDs to the annotations and language information they provide. 

\subsection{Dense Annotation Sampling}\label{sec:app:densesampling}

\begin{figure*}[t]
    \centering
    \includegraphics[width=\linewidth]{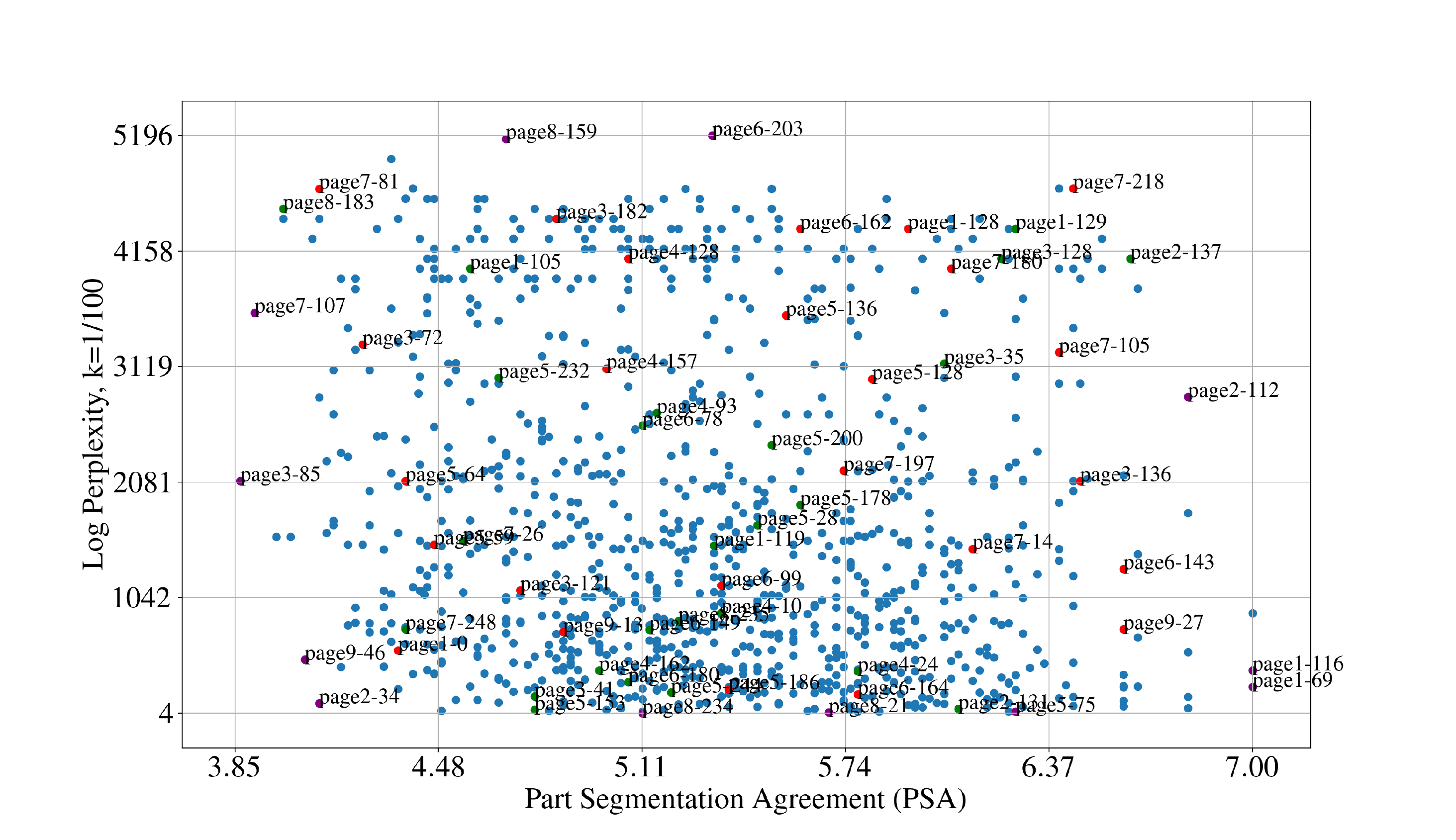}
    \caption{Sampled tangrams for dense annotation collection: 12 purple points picked from the periphery, 25 red points randomly sampled from each grid, and 25 green points uniformly sampled from all points.}\label{fig:sampling}
\end{figure*}

The set \kgdense is made of 62 tangrams sampled from \kgfull and 12 tangrams commonly used in prior work. 
We sample the 62 tangrams from \kgfull to represent the diversity of tangrams using the first set of annotations we collect. 
We  plot the annotated tangrams by average log perplexity of whole-shape descriptions with \( \frac{1}{100} \) smoothing and PSA and apply a 5 $\times$ 5 grid to the plot (Figure~\ref{fig:sampling}). Using perplexity and PSA allows us to sample a set of tangrams with diverse degrees of annotation and segmentation agreement. With a relatively high smoothing factor, we are able to spread out the data points, because the majority of the data set has high divergence in descriptions. We randomly pick 12 periphery points to collect more annotations for outliers, uniformly sample 25 from all the 1004 tangrams, and randomly sample 25, one from each grid, to represent the entire distribution. 

We calculate average log perplexity of whole-shape annotations for each tangram. 
Let $\tokenseq^{(1)},\dots,\tokenseq^{(N)}$ be annotations for a tangram, where each annotation is a sequence of tokens $\tokenseq^{(j)} = \langle \token_1,\dots,\token_{M^{(j)}}\rangle$ of length $M^{(j)}$. 
We create a language model $p^{(j)}$ for every annotation $\tokenseq^{(j)}$ using all other $N-1$ annotations for the tangram: 

\vspace{-5pt}
\begin{small}
\begin{equation}
    p^{(j)}(\token) = \frac{C_{\token \in \tokenseq^{(j'\neq j)}}+ k}{{\rm total}_{j'\neq j} + kV}\;\;,
\end{equation}
\end{small}
\vspace{-5pt}

\noindent
where $C_{\token \in \tokenseq^{(j'\neq j)}}$ is the number of occurrences of $\token$ in the other annotations for the tangram, $k$ is the smoothing factor, ${\rm total}_{j' \neq j}$ is the total number of words used in the other annotations for the tangram and $V$ is the vocabulary size of all whole-shape annotations across all tangrams. 
The log perplexity for annotation $\tokenseq^{(j)}$ is  $\log{PP^{(j)}}=-\frac{1}{M^{(j)}}\sum_{i=1}^{M^{(j)}}\log_2{p(\token^{(j)}_i)}$. The log perplexity for the tangram is the average of perplexity values for all its annotations  $\log{PP}=\frac{1}{N}\sum_{j=1}^N \log{PP^{(j)}}$.
We lowercase, stem, and remove stop words before computing the log perplexity.

\subsection{Example Inputs for Experimental Conditions}\label{sec:app:condexamples}

\begin{figure*}
    \centering
    \includegraphics[width=\textwidth]{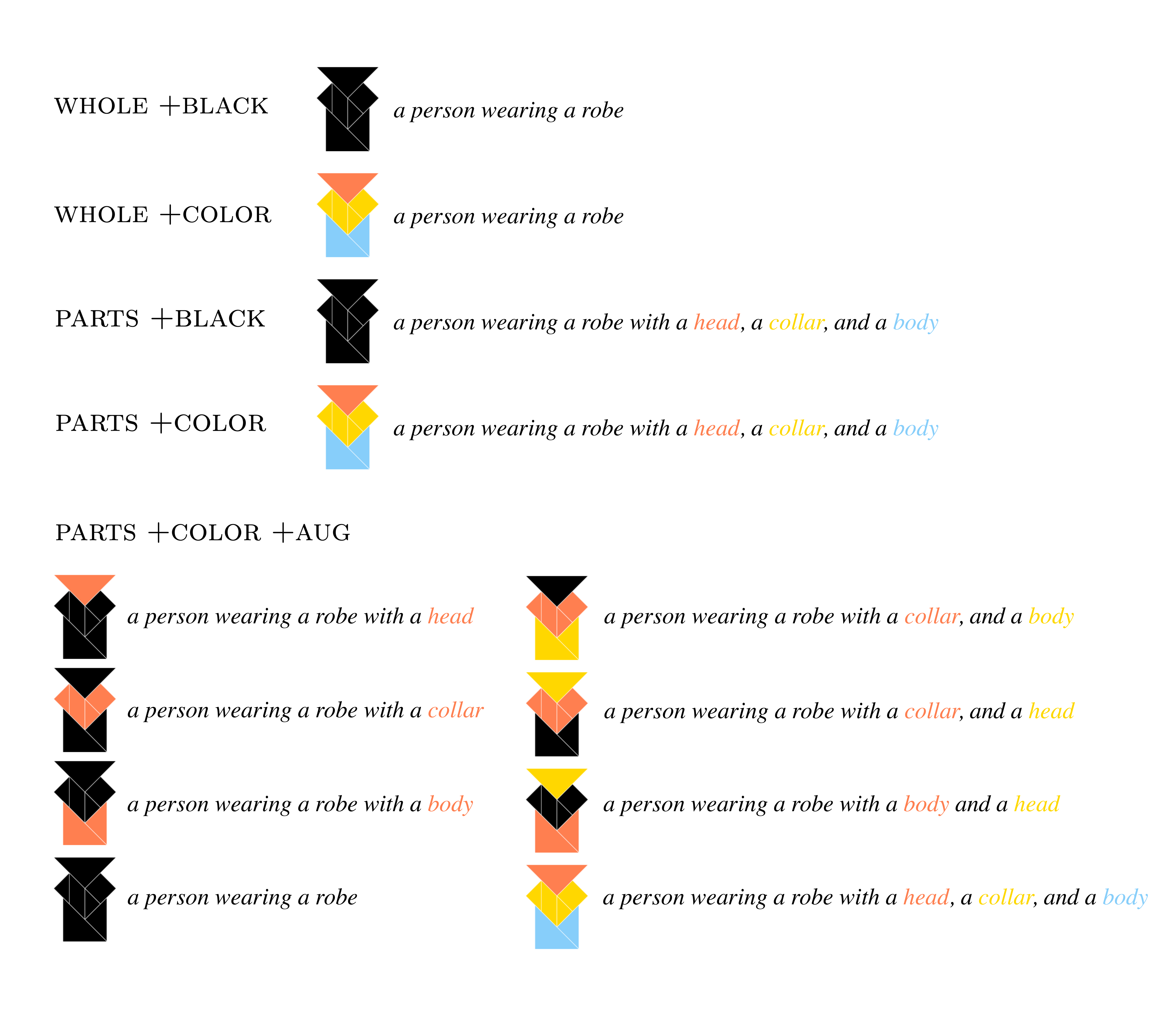}
    \caption{An example of one annotation across the different experimental conditions. The augmentation condition (\expcondaug) creates multiple examples from the same annotation.}
    \label{fig:tangramexamplesAnnotation}
\end{figure*}

\autoref{fig:tangramexamplesAnnotation} shows how one annotation, including both text and image, appears under the different experimental conditions. 
For conditions with \expcondparts\xspace annotations, we generate simple English sentences combining the whole shape description with part descriptions using the template \nlstring{<whole shape> with <part>, <part>, ..., and <part>}. We add an indefinite article to each singular part description. \expcondblack\xspace images are tangrams with all pieces colored black with white borders. \expcondcolored\xspace images are tangrams with each part colored with one of the CSS preset colors in the order of coral, gold, lightskyblue, lightpink, mediumseagreen, darkgrey, lightgrey that correspond to the parts in the annotation. 
For the augmented condition (\expcondaug), text inputs are whole annotations combined with each possible subset of the part descriptions. Image inputs are tangrams colored in the same way as colored images, but the parts excluded from the subset of part descriptions are colored black instead.
All part descriptions in the annotations are randomly shuffled and not consistently associated with any particular color in the images, so that the coloring solely serves as an indication of the ordering of parts in the combined text.

\subsection{Human Performance Baseline Details}\label{sec:app:human}

We recruited an independent group of 233 human  participants from the Prolific crowdsourcing platform (\url{https://www.prolific.co/}), and asked them to perform the same reference game task we used for model evaluation.
Each participant was randomly assigned to one of the four conditions and shown a random sequence of 20 trials from that condition.
On each trial, we showed a text annotation from the development set along with the corresponding context of ten tangrams and asked the participant to click the tangram that was being described.
The information that was available varied across condition, just as in the model evaluations. 
The tangrams were either presented to participants in black-and-white (\expcondblack) or colored according to their segmentation map (\expcondcolored), and the language was either the whole-shape description alone (\expcondwhole) or with the parts included (\expcondparts).
In the \expcondparts+\expcondcolored{} condition, the parts text was colored to match the image to facilitate visual comparison, providing the same alignment information available to the models.

We took several steps to ensure high-quality responses. 
First, participants began with a fixed set of 10 practice trials to familiarize with the task. 
For these practice trials, we provided feedback indicating whether they have selected the correct tangram, and if not, we highlight the correct answer.
To assess whether participants were paying attention as opposed to responding randomly, we inserted an unambiguous ``catch trial'' where the target was the square tangram and the description was \nlstring{square}. 
We excluded 16 participants who failed to select the correct target on this trial, yielding a final sample size of 217 participants out of the 233 recruited.
Because our aim was to obtain overall accuracy estimates for each condition, we did not require judgements for every individual annotation and context in the test set. 
However, we were able to ensure good coverage of the dataset, including annotations from all 125 tangrams and over 600 unique descriptions in each condition.

\subsection{Model-specific Implementation Details}\label{sec:app:implmodels}

For experiments with \clip, we use the ViT-B/32 variant. 
We fine-tune using an Adam optimizer with learning rate 5e-8 and weight decay 1e-6. At the end of each epoch, the training data is shuffled and rebatched. 
We train the models up to 200 epochs and use patience of 50 epochs to select the model with the highest image prediction accuracy on a non-augmented validation set taken from the training data.
All images are resized to \clip's default input resolution of 224 $\times$ 224, with white padding to make to rectangle images square.
The total number of trainable parameters in \clip is 151.2M. 
\clip models are fine-tuned with either a single GeForce RTX 2080 Ti GPU with 11GB memory or a single Titan RTX GPU with 24GB memory. 
Fine-tuning takes approximately 40 minutes per epoch for augmented setups (\expcondaug) and roughly 3 minutes for other setups.

For ViLT experiments, we fine-tune with an AdamW optimizer with learning rate 1e-4 and weight decay 1e-2. 
We use a cosine learning rate schedule with warm-up over the first epoch.
We train the models up to 30 epochs with a patience of 10 epochs and follow the same model selection criterion as for \clip.
All images are resized to 384 $\times$ 384. 
The total number of trainable parameters in \vilt is 87.4M.
\vilt models are fine-tuned with a single Titan RTX GPU with 24 GB memory.
Fine-tuning takes up to 5.5 hours per epoch for augmented setups (\expcondaug) and roughly 15 minutes for other setups.

\subsection{Random Generation of Reference Games}\label{sec:app:randomrefgames}

\begin{figure}[t!]
    \centering
    \footnotesize
    \includegraphics[width=\linewidth]{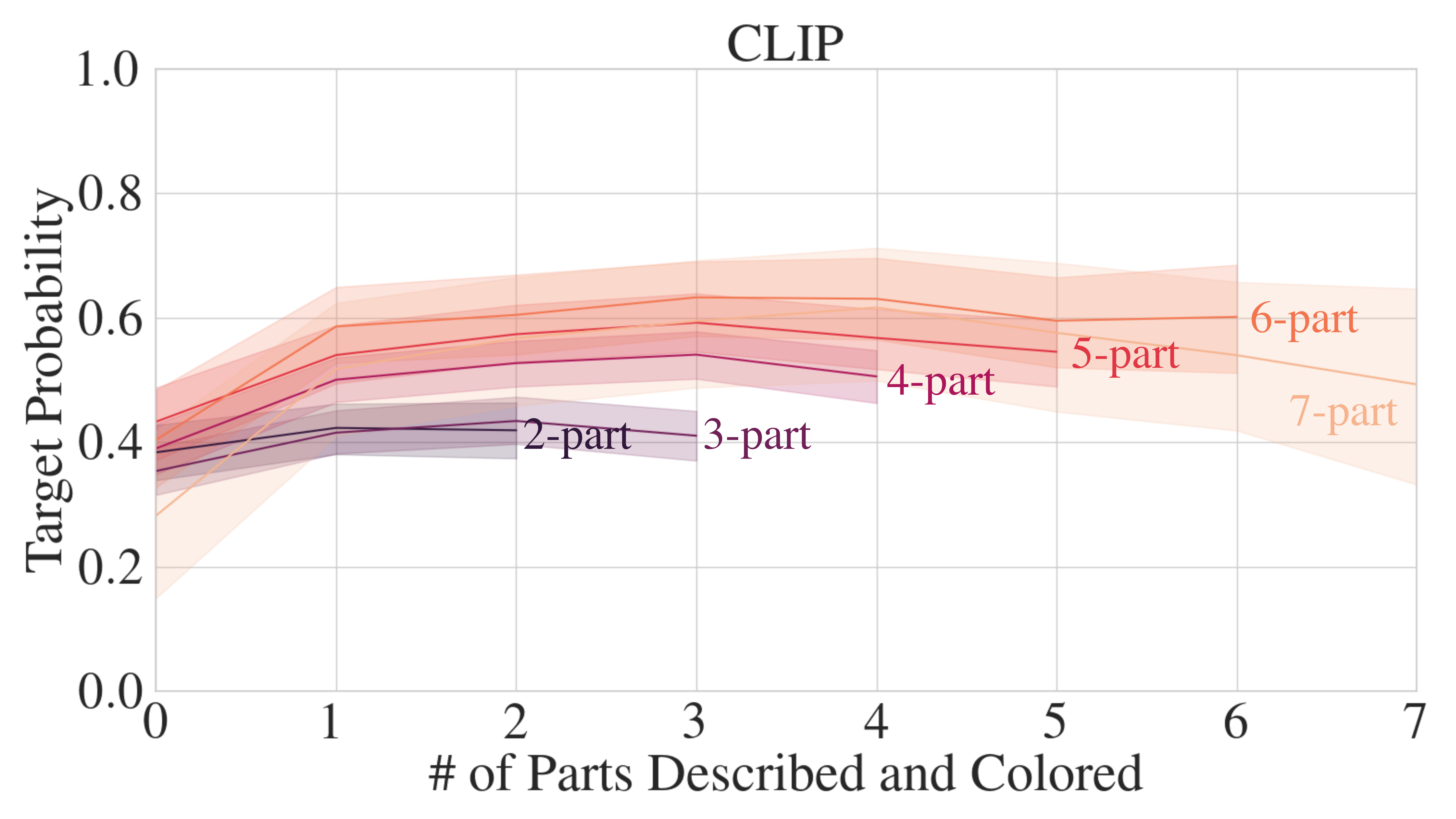}
    \medskip
    \includegraphics[width=\linewidth]{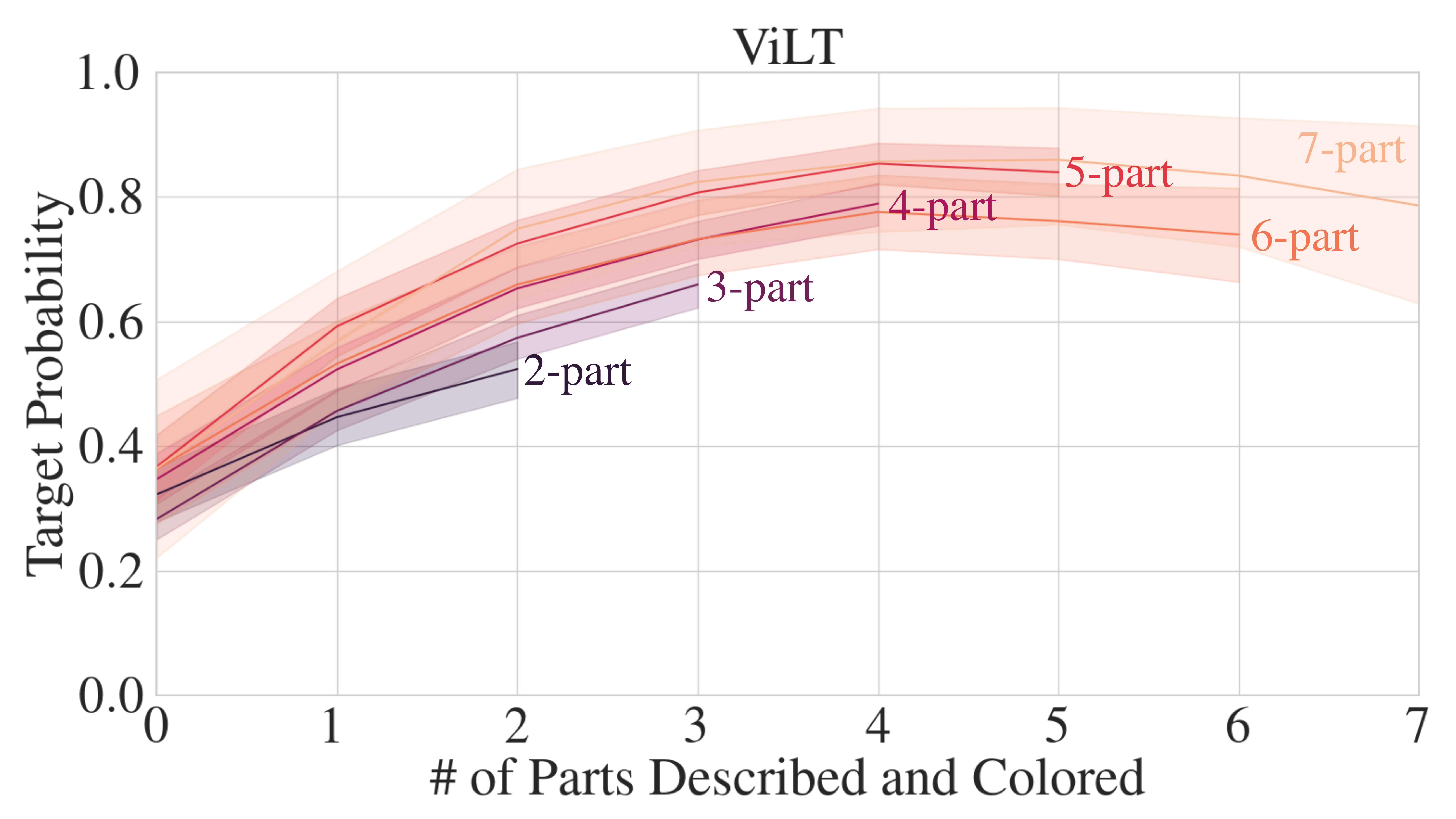}
    \caption{Mean development probabilities of predicting the correct image in reference games generated without constraints using fine-tuned \clip (top) or fine-tuned \vilt (bottom) by number of parts included in text and colored in the images. We separate the curves by the total number of parts in the annotation of the target example. The error bands show the 95\% confidence interval of the expected mean at each point by bootstrapping with 1000 resamplings.}
    \label{fig:randompartplots}
\end{figure}

In our main experiments (\autoref{sec:absvis}), we randomly generate reference games subject to constraints (\autoref{sec:absvis:gamegen}). 
In particular, we ensure that distractors contained the same total number of parts.
We explore the impact of these constraints by repeating our experiments on reference games generated without the constraints. 
Without the constraints, part counting can help the model disqualify distractors and significantly narrow down the set of likely referents. 
This is because images with a different number of parts colored compared to the number of parts in the text description can be easily ignored without considering the semantics of the text or images. 
\autoref{tab:randomperformance} shows development accuracies for games generated without constraints, both for training and testing. 
Generally, the success rate achieved on unconstrained contexts is much higher compared to contexts generated with constraints (\autoref{tab:performance}). 
However, when analyzing the performance of this model on part-controlled contexts (\autoref{fig:randompartplots}), we observe roughly similar performance to the games generated with constraints (\autoref{fig:partplots}), even though we would expect a significant performance increase given the results in \autoref{tab:randomperformance}.
We even observe a more pronounced decrease in performance when more parts are added, illustrating further difficulty generalizing. 
We conclude that the model trained on games generated without constraints (\autoref{tab:randomperformance}) likely learns to rely on part-counting heuristics and may be less reliable in other settings.

\begin{table}[t]
    \footnotesize\centering
    \begin{tabular}{@{}lcccc@{}}
        \toprule
        \multirow{2}{*}{\textbf{Condition}} & \multicolumn{2}{c}{\clip}  & \multicolumn{2}{c}{\vilt} \\
        & PT & FT & PT & FT \\
        \midrule
        \expcondwhole+\expcondblack & 17.3 & 46.2 & 13.2 & 41.3 \\
        \expcondparts+\expcondblack & 16.8 & 47.4 & 12.6 & 47.0 \\
        \expcondwhole+\expcondcolored & 15.9 & 48.0 & 12.4 & 46.2 \\
        \expcondparts+\expcondcolored & 15.9 & 71.3 & 12.1 & 89.0  \\
        \expcondparts+\expcondcolored+\expcondaug  & -- & 74.0 & -- & 86.0 \\
        \bottomrule
    \end{tabular}
    \caption{Reference game development accuracies (\%) for the different experimental conditions with pre-trained (PT) or fine-tuned (FT) models for games generated without constraints. }
    \label{tab:randomperformance}
\end{table}

\subsection{Reproducibility Checklist}\label{sec:app:reproduce}

\noindent For all reported experimental results:
\begin{itemize}
    \item A clear description of the mathematical setting, algorithm, and/or model: yes; see \autoref{sec:absvis}.
    \item Submission of a zip file containing source code, with specification of all dependencies, including external libraries, or a link to such resources: yes; attached to our submission.
    \item Description of computing infrastructure used: yes; see \aautoref{sec:app:implmodels}.
    \item The average runtime for each model or algorithm (e.g., training, inference, etc.) or estimated energy cost: yes; see \aautoref{sec:app:implmodels}.
    \item Number of parameters in each model: yes; see \aautoref{sec:app:implmodels}.
    \item Corresponding validation performance for each reported test result: yes; see \aautoref{tab:performance} and \aautoref{tab:randomperformance} for results on the development set.
    \item Explanation of evaluation metrics, with links to code used: yes; see \autoref{sec:absvis} for an explanation of the reference game metric. An implementation is included in the attached code zipfile.
\end{itemize}

For all experiments with hyperparameter search:

\begin{itemize}
    \item We performed a minimal manual search for learning rate and weight decay, and used the same values for all experiments (described in \autoref{sec:app:implmodels}).
\end{itemize}

For all datasets used:

\begin{itemize}
    \item Relevant details such as languages, and number of examples and label distributions: yes; see \autoref{sec:data}.
    \item Details of train/test/validation splits: yes; see \autoref{sec:data:split}.
    \item Explanation of any data that were excluded, and all pre-processing steps: yes; see \autoref{sec:data}  and \autoref{sec:app:scanning}.
    \item A zip file containing data or link to a downloadable version of the data: yes; attached to our submission.
    \item For new data collected, a complete description of the data collection process, such as instructions to annotators and methods for quality control: yes; see \autoref{sec:data:language} and \autoref{sec:app:crowdsourcing}.
\end{itemize}

\end{document}